\documentclass[lettersize,journal]{IEEEtran}
\usepackage{amsmath,amsfonts}
\usepackage{algorithmic}
\usepackage{algorithm}
\usepackage{array}
\usepackage{subcaption}
\usepackage{textcomp}
\usepackage{stfloats}
\usepackage{url}
\usepackage{verbatim}
\usepackage{graphicx}
\usepackage{svg}
\usepackage{hyperref}
\usepackage{cite}
\usepackage{marvosym}
\usepackage{multirow}
\usepackage{booktabs}
\usepackage{tikz}
\usetikzlibrary{arrows.meta, positioning, calc}
\usepackage{standalone} % 必须引入这个包
\usepackage{tikz}

\begin{document}

% \title{A Survey of Drift Learning: Paving the Path to Autonomous Learning Systems}
\title{Autonomous Drift Learning in Data Streams: A Unified Perspective}

\author{ Xiaoyu~Yang, 
         En Yu,
         Jie Lu \textsuperscript{(\Letter)}~\IEEEmembership{Fellow,~IEEE}        
        % <-this % stops a space
\thanks{Xiaoyu Yang, Jie Lu and En Yu are with the Australian Artificial Intelligence Institute (AAII), Faculty of Engineering and Information Technology, University of Technology Sydney, Australia}
\thanks{This work has been submitted to the IEEE for possible publication. Copyright may betransferred without notice, after which this version may no longer be accessible.}
}


\markboth{Journal of \LaTeX\ Class Files,~Vol.~14, No.~8, August~2021}%
{Shell \MakeLowercase{et al.}: A Sample Article Using IEEEtran.cls for IEEE Journals}


\maketitle

\begin{abstract}
In the pursuit of autonomous learning systems, the foundational assumption of stationarity, the premise that data distributions and model behaviors remain constant, is fundamentally untenable. Historically, the research community has addressed non-stationary environments almost exclusively under the scope of concept drift, focusing primarily on temporal shifts in streams. However, as learning systems become increasingly autonomous and complex, merely adapting to temporal non-stationarity is no longer sufficient. Evolving beyond this traditional perspective, we propose a novel, three-dimensional taxonomy that systematizes the field based on the operational state of the system. First, time stream drift distinguishes between stochastic arbitrary patterns and structural rhythmic dynamics. Second, data stream drift disentangles shifts in feature representations, identified as representation drift, from changes in underlying semantics, recognized as semantic drift. Third, model stream drift characterizes the internal endogenous divergence of learning systems through the lenses of sequential plasticity, decentralized heterogeneity, and policy instability. Based on this framework, we systematically review 193 representative studies and identify key open challenges. By bridging the fragmented paradigms of drift adaptation, continual learning, and temporal generalization, this survey outlines a roadmap for building self-evolving intelligent systems capable of learning autonomously through continuous change.
\end{abstract}

\begin{IEEEkeywords}
Concept Drift, Data Streams, Continual Learning
\end{IEEEkeywords}

\section{Introduction}

\IEEEPARstart{T}owards autonomous learning systems, the assumption of stationarity, the premise that data distributions and model behaviors remain constant over time, is increasingly untenable. Historically, these challenges have been primarily investigated under the scope of concept drift~\cite{gama2014survey,lu2018learning}, which characterizes the non-stationary evolution of data streams over temporal intervals. In this survey, we redefine this foundational temporal perspective as time stream drift and employ it as a logical cornerstone to expand our understanding toward a more comprehensive framework of drift learning. To provide a rigorous theoretical foundation, we establish a unified definition of drift based on the operational state of the system. We define the state at time $t$ as a tuple $\Omega_t = \langle P_t(X, Y), \theta_t \rangle$, where $P_t(X, Y)$ represents the joint probability distribution of the external data stream~\cite{lu2018learning} and $\theta_t \in \Theta$ denotes the internal hypothesis parameters or structural configuration of the system~\cite{heDYSONDynamicFeature2024}. Consequently, generalized drift is defined as a non-stationary deviation in the system operational state across temporal or spatial dimensions, expressed as $\Omega_t \neq \Omega_{t+\Delta t}$ or $\Omega^{(i)} \neq \Omega^{(j)}$.

This unified perspective enables the categorization of drift learning into three interconnected dimensions, expanding from familiar temporal dynamics to environmental content and internal system states. First, \textbf{Time Stream Drift} describes the temporal dynamics and evolution patterns of the state deviation $\Delta \Omega$. This dimension focuses on the mechanisms of change over time, distinguishing between stochastic patterns, referred to as arbitrary drift, and structural regularities, denoted as rhythmic drift~\cite{yu2025learning}. Second, \textbf{Data Stream Drift} characterizes the deviation of the external environment $P_t(X, Y)$, where $P_t(X, Y) \neq P_{t+\Delta t}(X, Y)$. It addresses variations in the data content itself, encompassing shifts in feature representations $P(X)$ and underlying semantics $P(Y|X)$. Third, \textbf{Model Stream Drift} focuses on the internal variation of the system state $\theta_t$. Even under stable data distributions, a model may experience endogenous divergence, denoted as $\theta_t \neq \theta_{t+\Delta t}$ or $\theta^{(i)} \neq \theta^{(j)}$, due to factors such as sequential forgetting in sequence drift, asynchronous updates in distributed networks leading to heterogeneity drift, or feedback-driven instability in decision-making through policy drift~\cite{caoDriftShieldAutonomousFraud2025,sunMOSModelSurgery2025}. Together, these three streams provide a holistic lens through which drift learning is analyzed as an integrated process of adaptation, consolidation, and evolution. This framework enables autonomous learning systems to learn through change by automatically identifying correlations and autonomously recalibrating upon the detection of drift.

Over the past decade, several surveys have systematically reviewed the progress of drift-related studies, primarily under the umbrella of concept drift. Early works such as Lu et al.~\cite{luLearningConceptDrift2019a} established the foundational taxonomy of drift detection and adaptation, inspiring subsequent studies that extended this framework to specific domains, including process mining~\cite{satoSurveyConceptDrift2021a}, data stream mining~\cite{agrahariConceptDriftDetection2022}, and regression tasks~\cite{limaLearningConceptDrift2022}. Later reviews refined the methodological spectrum by distinguishing active and passive strategies~\cite{hanSurveyActivePassive2022} or by linking drift detection to model degradation and performance monitoring~\cite{bayramConceptDriftModel2022a}. Despite these advances, existing reviews remain largely confined to data-level or algorithmic perspectives, often treating drift as an external perturbation to be detected and mitigated. They seldom address how drift propagates through model evolution, interacts with temporal dynamics, or manifests as an inherent process shaping long-term autonomy.

In contrast, this review aims to go beyond the conventional notion of concept drift and present a unified perspective of drift learning. We systematically organize the literature across three complementary axes—data stream drift, model stream drift, and time stream drift—to uncover how data, model, and temporal dynamics jointly drive adaptive behavior. By bridging fragmented paradigms from drift detection, continual learning, and temporal adaptation, our work highlights drift learning as a foundational principle toward building self-evolving, autonomous learning systems.

To ensure a comprehensive and unbiased coverage of existing studies, we adopted a structured literature selection process following systematic review principles. The procedure consisted of four main stages:

\begin{figure*}[htbp]
    \centering
    \includegraphics[width=0.95\textwidth]{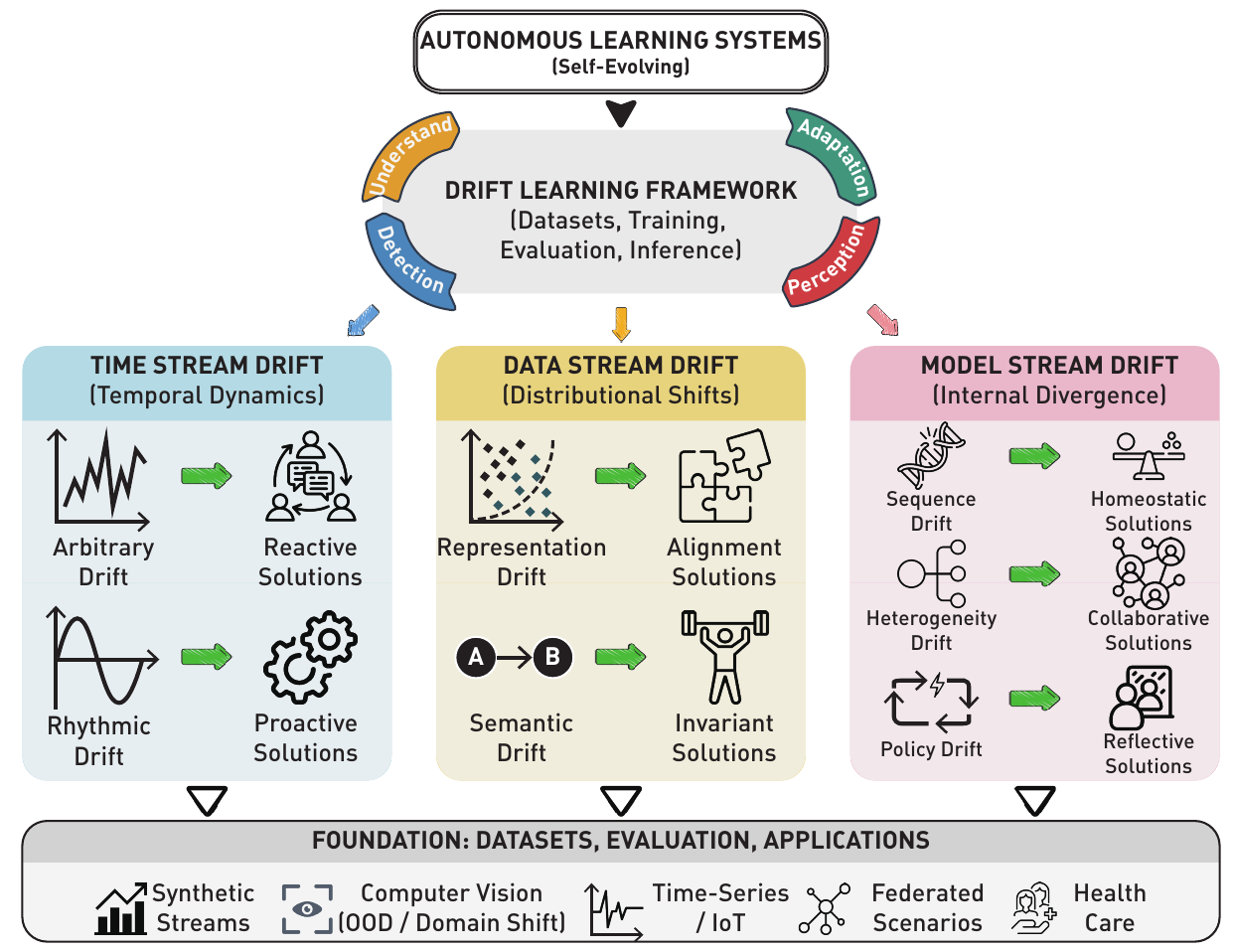}
    \caption{An overview of the proposed drift learning framework. This framework categorizes drift phenomena into three interconnected dimensions: time stream drift (addressing temporal dynamics), data stream drift (handling data-generating processes), and model stream drift (managing internal model evolution). By integrating adaptation, consolidation, and evolution mechanisms supported by comprehensive benchmarks, the framework aims to endow autonomous learning systems with the ability to perceive, adapt to, and anticipate changes in open environments through automated feature alignment and self-sustained training loops.}
    \label{fig:placeholder}
\end{figure*}

\begin{enumerate}
    \item Database Selection

        We collected publications from major scientific databases including IEEE Xplore, ACM Digital Library, ScienceDirect, SpringerLink, which jointly cover most high-impact venues in knowledge engineering, data mining, and machine learning.

    \item Keyword Strategy

        The search queries combined general and specific terms related to drift and adaptation, such as “concept drift,” “model drift,” “data stream adaptation,” “temporal drift,” “continual learning,” “non-stationary environments,” and “autonomous learning systems.” Boolean operators and truncations (e.g., “drift*” OR “adapt*”) were applied to capture terminological variations across disciplines.

    \item Inclusion and Exclusion Criteria

        Inclusion: peer-reviewed journal or top-tier conference papers (2015–2025) directly addressing drift detection, adaptation, or learning under non-stationarity; studies that explicitly model data, model, or temporal evolution.

        Exclusion: works focusing solely on static domain adaptation, incremental dataset expansion without drift context, or papers lacking methodological or theoretical contributions.

    \item Screening and Consolidation

        After initial retrieval, all titles and abstracts were screened for relevance. Duplicates were removed, and the remaining papers were manually verified for alignment with at least one of the three analytical axes, data stream drift, model stream drift, or time stream drift. The final corpus consisted of both foundational studies and recent advances, complemented by representative survey articles such as~\cite{luLearningConceptDrift2019a, satoSurveyConceptDrift2021a, agrahariConceptDriftDetection2022, bayramConceptDriftModel2022a}.
    
\end{enumerate}

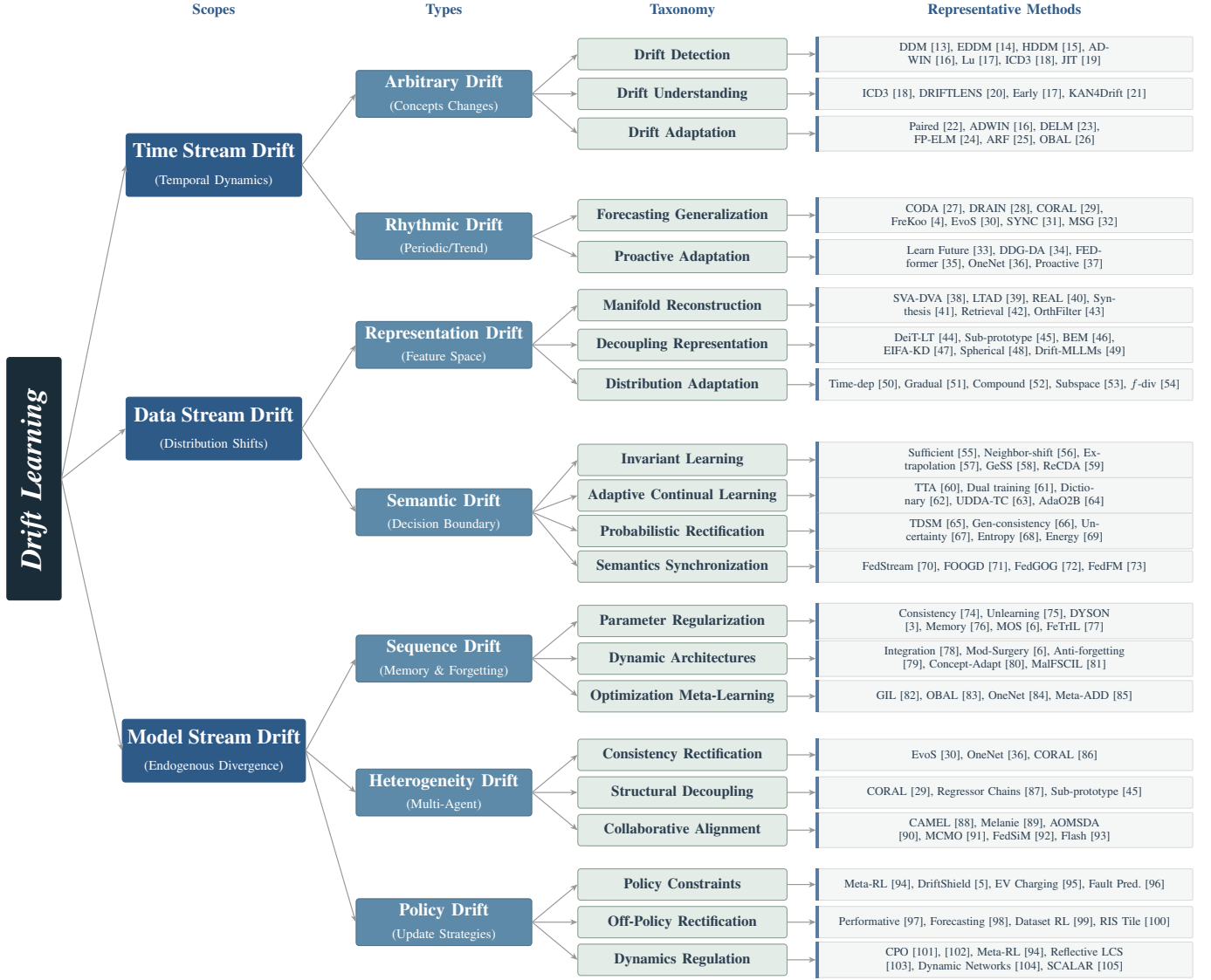
\begin{figure*}[t]
  \centering
  \resizebox{\textwidth}{!}{
    \begin{tikzpicture}
      %% ── 颜色定义 ──────────────────────────────────────────────────────
      \definecolor{colorRoot}{HTML}{1A2C38}
      \definecolor{colorScope}{HTML}{2E5A88}
      \definecolor{colorType}{HTML}{5D8AA8}
      \definecolor{colorMeth}{HTML}{A3C1AD}
      \definecolor{colorList}{HTML}{F4F7F6}
      \definecolor{mtext}{HTML}{2C3E50}

      %% ── 样式定义 ──────────────────────────────────────────────────────
      \tikzset{
        root/.style={
          draw=colorRoot, fill=colorRoot, text=white,
          rounded corners=2pt, minimum width=5.8cm, minimum height=1.3cm,
          font=\huge\bfseries\itshape, align=center
        },
        scope/.style={
          draw=colorScope, fill=colorScope, text=white,
          rounded corners=2pt, minimum width=4.2cm, minimum height=1.5cm,
          font=\Large\bfseries, align=center
        },
        type/.style={
          draw=colorType!80!black, fill=colorType, text=white,
          rounded corners=2pt, minimum width=4.2cm, minimum height=1.1cm,
          font=\large\bfseries, align=center
        },
        meth/.style={
          draw=colorMeth!80!black, fill=colorMeth!30, text=mtext,
          rounded corners=2pt, minimum width=5cm, minimum height=0.75cm,
          font=\bfseries, align=center
        },
        replnode/.style={
          draw=gray!30, fill=colorList, text=mtext,
          rounded corners=1pt, minimum width=9.0cm, minimum height=0.75cm, text width=8.7cm,
          font=\footnotesize, align=center, inner sep=4pt, line width=0.5pt
        },
        arr/.style={-{Stealth[length=5pt,width=4pt]}, gray!80, line width=0.7pt},
      }

      %% ── 辅助宏 ────────────────────────────────
      \newcommand{\methodnode}[4]{
        \node[replnode] (#1) at (#2,#3) {#4};
        \fill[colorScope!80] 
          ($(#1.north west)$) -- ($(#1.north west)+(0.08,0)$) -- 
          ($(#1.south west)+(0.08,0)$) -- ($(#1.south west)$) -- cycle;
      }

      %% ── 坐标定义 ──────────────────────────────────────────────────────
      \def\XRoot{0} \def\XScope{4.3} \def\XType{9.8} \def\XMeth{15.5} \def\XScen{23.2}

      %% ── 绘图内容 ──────────────────────────────────────────────────────
      \node[root, rotate=90] (root) at (\XRoot,0) {Drift Learning};

      %% SCOPE 1
      \node[scope] (S1) at (\XScope, 7.5) {Time Stream Drift\\[1pt]{\normalfont\small(Temporal Dynamics)}};
      \node[type] (T11) at (\XType, 9.2) {Arbitrary Drift\\[1pt]{\normalfont\small(Concepts Changes)}};
      \node[meth] (M111) at (\XMeth, 10.15) {Drift Detection};
      \node[meth] (M112) at (\XMeth, 9.20) {Drift Understanding};
      \node[meth] (M113) at (\XMeth, 8.25) {Drift Adaptation};
      \methodnode{C111}{\XScen}{10.15}{DDM \cite{gama2004learning}, EDDM \cite{baena2006early}, HDDM \cite{frias2014online}, ADWIN \cite{bifet2007learning}, Lu \cite{lu2025early}, ICD3 \cite{zhang2025learning}, JIT \cite{alippi2008just}}
      \methodnode{C112}{\XScen}{9.20}{ICD3 \cite{zhang2025learning}, DRIFTLENS \cite{greco2025unsupervised}, Early \cite{lu2025early}, KAN4Drift \cite{xu2024kan4drift}}
      \methodnode{C113}{\XScen}{8.25}{Paired \cite{bach2008paired}, ADWIN \cite{bifet2007learning}, DELM \cite{xu2017dynamic}, FP-ELM \cite{liu2016fp}, ARF \cite{gomes2017adaptive}, OBAL \cite{yu2024online}}

      \node[type] (T12) at (\XType, 5.8) {Rhythmic Drift\\[1pt]{\normalfont\small(Periodic/Trend)}};
      \node[meth] (M121) at (\XMeth, 6.3) {Forecasting Generalization};
      \node[meth] (M122) at (\XMeth, 5.3) {Proactive Adaptation};
      \methodnode{C121}{\XScen}{6.3}{CODA \cite{chang2023coda}, DRAIN \cite{nasery2021training}, CORAL \cite{xu2025drift2matrix}, FreKoo \cite{yu2025learning}, EvoS \cite{xie2024evolving}, SYNC \cite{he2025learning}, MSG \cite{yu2026autonomous}}
      \methodnode{C122}{\XScen}{5.3}{Learn Future \cite{you2021learning}, DDG-DA \cite{li2022ddg}, FEDformer \cite{zhou2022fedformer}, OneNet \cite{wen2023onenet}, Proactive \cite{zhao2025proactive}}

      %% SCOPE 2
      \node[scope] (S2) at (\XScope, 1.2) {Data Stream Drift\\[1pt]{\normalfont\small(Distribution Shifts)}};
      \node[type] (T21) at (\XType, 3.2) {Representation Drift\\[1pt]{\normalfont\small(Feature Space)}};
      \node[meth] (M211) at (\XMeth, 4.15) {Manifold Reconstruction};
      \node[meth] (M212) at (\XMeth, 3.20) {Decoupling Representation};
      \node[meth] (M213) at (\XMeth, 2.25) {Distribution Adaptation};
      \methodnode{C211}{\XScen}{4.15}{SVA-DVA \cite{chenDelvingTrajectoryLongtail2024}, LTAD \cite{hoLongTailedAnomalyDetection2024}, REAL \cite{parasharNeglectedTailsVisionLanguage2024}, Synthesis \cite{liSynthesizingMinoritySamples2025}, Retrieval \cite{sidhuSearchDetectTrainingFree2025}, OrthFilter \cite{young2025fewer}}
      \methodnode{C212}{\XScen}{3.20}{DeiT-LT \cite{rangwaniDeiTLTDistillationStrikes2024}, Sub-prototype \cite{wangLongTailClassIncremental2024}, BEM \cite{zhengBEMBalancedEntropybased2024}, EIFA-KD \cite{dengEIFAKDExplicitImplicit2026}, Spherical \cite{yangTdistributedSphericalFeature2023}, Drift-MLLMs\cite{yangAdaptingMultimodalLarge2024b}}
      \methodnode{C213}{\XScen}{2.25}{Time-dep \cite{heDomainAdaptationTime2023}, Gradual \cite{saberiGradualDomainAdaptation2024}, Compound \cite{fengOpenCompoundDomain2023}, Subspace \cite{liSubspaceIdentificationMultiSource2023}, $f$-div \cite{wangFDivergencePrincipledDomain2024}}

      \node[type] (T22) at (\XType, -0.8) {Semantic Drift\\[1pt]{\normalfont\small(Decision Boundary)}};
      \node[meth] (M221) at (\XMeth, 0.45) {Invariant Learning};
      \node[meth] (M222) at (\XMeth, -0.4) {Adaptive Continual Learning};
      \node[meth] (M223) at (\XMeth, -1.25) {Probabilistic Rectification};
      \node[meth] (M224) at (\XMeth, -2.1) {Semantics Synchronization};
      \methodnode{C221}{\XScen}{0.45}{Sufficient \cite{kimSufficientInvariantLearning2025}, Neighbor-shift \cite{liLetInvariantLearning2025}, Extrapolation \cite{liGraphStructureExtrapolation2024}, GeSS \cite{zouGeSSBenchmarkingGeometric2024}, ReCDA \cite{yangReCDAConceptDrift2024}}
      \methodnode{C222}{\XScen}{-0.4}{TTA \cite{kimTestTimeAdaptationInduces2024}, Dual training \cite{yangDualTestTimeTraining2025}, Dictionary \cite{yangOODDTesttimeOutofDistribution2025}, UDDA-TC \cite{liuUDDATCUnsupervisedRealTime2025}, AdaO2B \cite{zhangAdaO2BAdaptiveOnline2025}}
      \methodnode{C223}{\XScen}{-1.25}{TDSM \cite{naLabelNoiseRobustDiffusion2023}, Gen-consistency \cite{shenOptimizingOODDetection2024}, Uncertainty \cite{mehrtensBenchmarkingCommonUncertainty2023}, Entropy \cite{neoMaxEntLossConstrained2024}, Energy \cite{tianModelingDistributionalUncertainty2023}}
      \methodnode{C224}{\XScen}{-2.1}{FedStream \cite{mawuliFedStreamPrototypeBasedFederated2023}, FOOGD \cite{liaoFOOGDFederatedCollaboration2024}, FedGOG \cite{zhouFedGOGFederatedGraph2025}, FedFM \cite{zhaoFedFMFederatedFewshot2025}}

      %% SCOPE 3
      \node[scope] (S3) at (\XScope, -6.5) {Model Stream Drift\\[1pt]{\normalfont\small(Endogenous Divergence)}};
      \node[type] (T31) at (\XType, -4.3) {Sequence Drift\\[1pt]{\normalfont\small(Memory \& Forgetting)}};
      \node[meth] (M311) at (\XMeth, -3.4) {Parameter Regularization};
      \node[meth] (M312) at (\XMeth, -4.3) {Dynamic Architectures};
      \node[meth] (M313) at (\XMeth, -5.2) {Optimization Meta-Learning};
      \methodnode{C311}{\XScen}{-3.4}{Consistency \cite{amadorcoelhoConceptDriftDetection2023}, Unlearning \cite{arteltUnsupervisedUnlearningConcept2023}, DYSON \cite{heDYSONDynamicFeature2024}, Memory \cite{sunClassIncrementalLearning2023}, MOS \cite{sunMOSModelSurgery2025}, FeTrIL \cite{petitFeTrILFeatureTranslation2023}}
      \methodnode{C312}{\XScen}{-4.3}{Integration \cite{liDynamicIntegrationTaskSpecific2025a}, Mod-Surgery \cite{sunMOSModelSurgery2025}, Anti-forgetting \cite{sunAntiforgettingIncrementalLearning2024}, Concept-Adapt \cite{liConceptDriftAdaptation2024a}, MalFSCIL \cite{chaiMalFSCILFewShotClassIncremental2025}}
      \methodnode{C313}{\XScen}{-5.2}{GIL \cite{yuGeneralizedIncrementalLearning2025a}, OBAL \cite{yuOnlineBoostingAdaptive2024a}, OneNet \cite{zhangOneNetEnhancingTime2023a}, Meta-ADD \cite{yu2022meta}}

      \node[type] (T32) at (\XType, -7.5) {Heterogeneity Drift\\[1pt]{\normalfont\small(Multi-Agent)}};
      \node[meth] (M321) at (\XMeth, -6.6) {Consistency Rectification};
      \node[meth] (M322) at (\XMeth, -7.5) {Structural Decoupling};
      \node[meth] (M323) at (\XMeth, -8.4) {Collaborative Alignment};
      \methodnode{C321}{\XScen}{-6.6}{EvoS \cite{xie2024evolving}, OneNet \cite{wen2023onenet}, CORAL \cite{xu2025coral}}
      \methodnode{C322}{\XScen}{-7.5}{CORAL \cite{xu2025drift2matrix}, Regressor Chains \cite{zhang2025tracking}, Sub-prototype \cite{wangLongTailClassIncremental2024}}
      \methodnode{C323}{\XScen}{-8.4}{CAMEL \cite{yu2025drift}, Melanie \cite{du2019multi}, AOMSDA \cite{renchunzi2022automatic}, MCMO \cite{jiao2022reduced}, FedSiM \cite{wangFedSiMSimilarityMetric2023}, Flash \cite{panchalFlashConceptDrift2023b}} 

      \node[type] (T33) at (\XType, -10.6) {Policy Drift\\[1pt]{\normalfont\small(Update Strategies)}};
      \node[meth] (M331) at (\XMeth, -9.7) {Policy Constraints};
      \node[meth] (M332) at (\XMeth, -10.6) {Off-Policy Rectification};
      \node[meth] (M333) at (\XMeth, -11.5) {Dynamics Regulation};
      \methodnode{C331}{\XScen}{-9.7}{Meta-RL \cite{ajayDistributionallyAdaptiveMeta2022}, DriftShield \cite{caoDriftShieldAutonomousFraud2025}, EV Charging \cite{poddubnyyOnlineEVCharging2023}, Fault Pred. \cite{shayestehAutomatedConceptDrift2022a}}
      \methodnode{C332}{\XScen}{-10.6}{Performative \cite{perdomoPerformativePrediction2020}, Forecasting \cite{zhaoPerformativeTimeSeriesForecasting2025}, Dataset RL \cite{schweighoferDatasetPerspectiveOffline2022}, RIS Tile \cite{wuGeneralizedRISTile2024}}
      \methodnode{C333}{\XScen}{-11.5}{CPO \cite{yang2025walking,yang2026towards}, Meta-RL \cite{ajayDistributionallyAdaptiveMeta2022}, Reflective LCS \cite{steinReflectiveLearningClassifier2021}, Dynamic Networks \cite{maniasModelDriftDynamic2023}, SCALAR\cite{yang2026scalar}}

      %% 连接线
      \begin{scope}
        \draw[arr] (root.south) -- (S1.west); \draw[arr] (root.south) -- (S2.west); \draw[arr] (root.south) -- (S3.west);
        \draw[arr] (S1.east) -- (T11.west); \draw[arr] (S1.east) -- (T12.west);
        \draw[arr] (S2.east) -- (T21.west); \draw[arr] (S2.east) -- (T22.west);
        \draw[arr] (S3.east) -- (T31.west); \draw[arr] (S3.east) -- (T32.west); \draw[arr] (S3.east) -- (T33.west);
        \draw[arr] (T11.east) -- (M111.west); \draw[arr] (T11.east) -- (M112.west); \draw[arr] (T11.east) -- (M113.west);
        \draw[arr] (T12.east) -- (M121.west); \draw[arr] (T12.east) -- (M122.west);
        \draw[arr] (T21.east) -- (M211.west); \draw[arr] (T21.east) -- (M212.west); \draw[arr] (T21.east) -- (M213.west);
        \draw[arr] (T22.east) -- (M221.west); \draw[arr] (T22.east) -- (M222.west); \draw[arr] (T22.east) -- (M223.west); \draw[arr] (T22.east) -- (M224.west);
        \draw[arr] (T31.east) -- (M311.west); \draw[arr] (T31.east) -- (M312.west); \draw[arr] (T31.east) -- (M313.west);
        \draw[arr] (T32.east) -- (M321.west); \draw[arr] (T32.east) -- (M322.west); \draw[arr] (T32.east) -- (M323.west);
        \draw[arr] (T33.east) -- (M331.west); \draw[arr] (T33.east) -- (M332.west); \draw[arr] (T33.east) -- (M333.west);
        \draw[arr] (M111.east) -- (C111.west); \draw[arr] (M112.east) -- (C112.west); \draw[arr] (M113.east) -- (C113.west);
        \draw[arr] (M121.east) -- (C121.west); \draw[arr] (M122.east) -- (C122.west);
        \draw[arr] (M211.east) -- (C211.west); \draw[arr] (M212.east) -- (C212.west); \draw[arr] (M213.east) -- (C213.west);
        \draw[arr] (M221.east) -- (C221.west); \draw[arr] (M222.east) -- (C222.west); \draw[arr] (M223.east) -- (C223.west); \draw[arr] (M224.east) -- (C224.west);
        \draw[arr] (M311.east) -- (C311.west); \draw[arr] (M312.east) -- (C312.west); \draw[arr] (M313.east) -- (C313.west);
        \draw[arr] (M321.east) -- (C321.west); \draw[arr] (M322.east) -- (C322.west); \draw[arr] (M323.east) -- (C323.west);
        \draw[arr] (M331.east) -- (C331.west); \draw[arr] (M332.east) -- (C332.west); \draw[arr] (M333.east) -- (C333.west);
      \end{scope}

      %% 表头
      \node[font=\bfseries, colorScope] at (\XScope, 11.2) {Scopes};
      \node[font=\bfseries, colorScope] at (\XType,  11.2) {Types};
      \node[font=\bfseries, colorScope] at (\XMeth,  11.2) {Taxonomy};
      \node[font=\bfseries, colorScope] at (\XScen,  11.2) {Representative Methods};

    \end{tikzpicture}
  }
    \caption{The hierarchical taxonomy structure of Drift Learning proposed in this survey. We categorize the field along three axes: (1) Time Stream, distinguishing between Arbitrary and Rhythmic drifts; (2) Data Stream, covering Representation and Semantic drifts; and (3) Model Stream, addressing Sequence, Heterogeneity, and Policy drifts. The rightmost column summarizes the representative scenarios and paradigms for each category.}
    \label{fig:cases}
\end{figure*}

\section{Time Stream Drift}

\subsection{Preliminary}

\noindent\textbf{Definition.}
Time stream drift, commonly referred to as concept drift, denotes changes in the underlying data distribution of a streaming process over time~\cite{lu2018learning,gama2014survey}. Formally, consider a data stream $S_{0,t} = \left \{(X_{0,0}, y_{0,0}),...,(X_{i,t}, y_{i,t}), \right\}$ over the time interval $[0, t]$, where $X_{i,t}$ represents the features and $y_{i,t}$ is the associated label, concept drift occurs at $t+1$ if joint distribution $p_{t+1}(X, y) \neq p_{t}(X, y)$. In essence, when $p(y|X)$ changes, the decision boundary of the model becomes outdated, reducing its predictive performance in the new environment. Traditionally, concept drift is divided into two fundamental categories: real drift, where $p(y|X)$ changes and thus affects the decision function, and virtual drift, where only $p(X)$ changes while $p(y|X)$ remains stable, typically resulting from covariate shifts or changes in data sampling processes.

\noindent
Drift Types and Taxonomy.
The classical taxonomy categorizes concept drift into four temporal patterns: sudden, incremental, gradual, and reoccurring~\cite{lu2018learning}. However, as data sources become increasingly intertwined with complex temporal dependencies, this categorization becomes insufficient to characterize structured or predictable patterns. Motivated by phenomena like seasonal variations and behavioral cycles, we introduce a refined taxonomy that distinguishes between arbitrary drift and rhythmic drift, offering a temporally grounded perspective on distribution evolution.

Arbitrary drift refers to unpredictable, stochastic changes in the data distribution. These shifts often arise from spontaneous external perturbations, abrupt system reconfigurations, or unexpected events. Drifts in this category can occur at any time, with unknown onset and unknown evolution patterns~\cite{gama2004learning,bifet2007learning}. Arbitrary drift thus represents the most challenging regime for adaptive learning systems, encompassing sudden concept switches, irregular distribution shifts, and unstructured gradual transitions.

Rhythmic drift describes temporally structured and predictable changes that exhibit recurrent or quasi-periodic patterns. These drifts are often driven by intrinsic temporal regularities in the data source, such as user habits, seasonal behaviors, or cyclical sensor conditions~\cite{nasery2021training,yu2025learning}. Incremental and reoccurring drifts naturally fall under this category, as they reveal predictable patterns either through smooth continuous evolution or through the repetition of previously seen concepts. The recognition of rhythmic drift provides opportunities for proactive adaptation—models can anticipate distribution changes based on learned temporal dependencies, rather than merely reacting to observed discrepancies.

This refined taxonomy extends the traditional notion of concept drift by incorporating a temporal–structural lens that distinguishes between stochastic and rhythmic dynamics. It provides a more faithful description of distribution evolution in complex non-stationary environments and lays the conceptual groundwork for designing drift learning algorithms that effectively balance robustness to arbitrary drifts with the ability to anticipate and exploit rhythmic ones.

\begin{figure}
    \centering
    \includegraphics[width=1\linewidth]{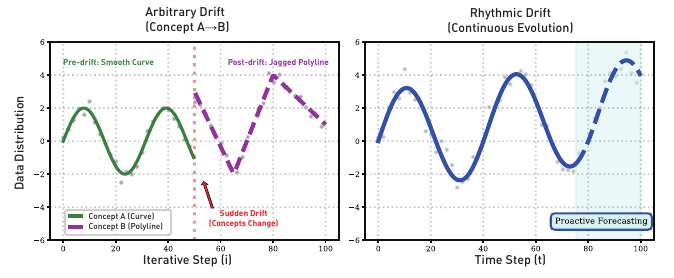}
    \caption{Visual comparison of the two sub-categories of time stream drift. \textbf{Left.} Arbitrary Drift, characterized by stochastic, unpredictable shifts, necessitating reactive detection mechanisms. \textbf{Right.} Rhythmic Drift: Exhibits temporal structures such as periodicity or smooth trends, enabling proactive forecasting and model anticipation}
    \label{fig:timeflow}
\end{figure}

\subsection{Arbitrary Drift}

Arbitrary drift characterizes stochastic and unpredictable distributional shifts within non-stationary environments. This phenomenon fundamentally violates the independent and identically distributed (i.i.d.) assumption central to static learning paradigms~\cite{hu2020domain,yu2025learning}. In streaming contexts, as the underlying data-generating process evolves continuously, models relying on historical priors often suffer severe performance deterioration when generalizing to newly arriving samples from shifted distributions. Consequently, timely and principled adaptation is critical for sustaining decision-making reliability in autonomous learning systems.

Formally, drift learning under arbitrary drift operates as a continuous loop comprising prediction, detection, and adaptation~\cite{lu2018learning}. At time step $t$, given an instance $X_t$,  the current hypothesis $H_t$ generates a prediction $\hat{y}_t$. Upon receiving the ground truth $y_t$ (or a delayed supervision signal), a detection mechanism evaluates whether a significant distribution shift (i.e., $P_{t+1} \neq P_t$) has occurred. 
If drift is identified, the adaptation phase transitions $H_t$ to $H_{t+1}$via strategies such as retraining, ensemble reweighting, or parameter tuning—to realign the learner with the emerging concept while preventing catastrophic forgetting. In essence, managing arbitrary drift requires a synergistic interplay between detection and adaptation, ensuring systems remain robust and efficient amidst highly volatile dynamics.

\subsubsection{Drift Detection}

Drift detection serves as the sensing mechanism for adaptive learning in streaming environments, identifying when and where distribution shifts occur. Historically, early approaches in both single and ensemble detection (e.g., DDM~\cite{gama2004learning}, EDDM~\cite{baena2006early}, HDDM~\cite{frias2014online}, ADWIN~\cite{bifet2007learning}, VFDT~\cite{domingos2000mining}) were predominantly traditional error-statistics or adaptive window based methods. While foundational, these classical statistical tests often struggle with the complexity of modern, high-dimensional streams. 

Consequently, recent advancements shift the focus toward leveraging deep features and uncertainty-based indicators to enhance robustness. In supervised settings, Lu et al.~\cite{lu2025early} propose a predictive-uncertainty index for early drift detection, supported by autonomous strategies~\cite{lu2025autonomous} that automatically recalibrate detection thresholds based on downstream performance. For unsupervised scenarios where labels are delayed, modern approaches directly measure distributional discrepancies in deep representation spaces. For instance, DRIFTLENS~\cite{greco2025unsupervised} monitors deep feature spaces, while ICD3~\cite{zhang2025learning} utilizes an Imbalanced Cluster Descriptor to maintain robust detection against concept imbalance. To filter spurious alarms, these deep indicators are increasingly integrated into multi-stage hypothesis testing frameworks~\cite{yu2017concept}, decoupling drift generation from validation.

\subsubsection{Drift Adaptation}

Drift adaptation represents the corrective phase that realigns the model's hypothesis with evolving distributions, navigating the stability-plasticity dilemma. Previous literature typically categorized adaptation into model replacement, ensemble management, and incremental adjustment. However, classical methods in these categories—such as maintaining paired learners~\cite{bach2008paired}, utilizing dynamically weighted ensembles (e.g., DWM~\cite{kolter2007dynamic}), or integrating drift detectors within tree structures like Adaptive Random Forests (ARF)~\cite{gomes2017adaptive}—often rely on cumbersome retraining or localized heuristics that struggle in complex, high-velocity environments.

To address these limitations, recent paradigms emphasize cross-stream correlation and test-time adaptation. In multi-source scenarios, adaptation has moved from simple ensemble weighting to explicit correlation modeling; for example, OBAL~\cite{yu2024online} tracks inter-stream correlations via adaptive boosting, which significantly enhances stability across concurrent, complex drifts. Concurrently, incremental adjustment has evolved into Test-Time Adaptation (TTA) and online recalibration mechanisms. By optimizing directly during the inference phase, these modern approaches allow deep learning systems to continuously align with streaming data on-the-fly, bypassing the need to store full historical trajectories or undergo costly model replacement.

\subsubsection{Drift Learning for Multiple Streams}
Most drift learning methods concentrate on single-stream settings. However, in real-world applications, data streams are often produced by multiple, concurrently evolving and non-stationary processes. For instance, real-time sentiment prediction on social media platforms such as Twitter. Incoming posts typically arrive without ground-truth sentiment labels, and obtaining annotations requires user feedback—which many users may be unwilling or unable to provide for privacy or personal reasons. As a result, building an accurate sentiment prediction model under limited or sparse supervision becomes challenging, necessitating new problem formulations and adaptive strategies. Motivated by the prevalence of such scenarios, recent works on multistream learning can be broadly categorized into two lines: i) Multistream classification, which leverages relationships across multiple evolving streams for predictive modeling; and ii) Multistream collaborative prediction, which aims to jointly exploit complementary information across streams to improve performance under drift and label scarcity.

i) Multistream classification aims to transfer knowledge from labeled source streams to unlabeled targets. While early solutions focused on density ratio estimation~\cite{haque2017fusion} and autonomous transfer learning~\cite{pratama2019atl}, recent advances leverage meta-learning frameworks~\cite{yu2022meta,yu2022learn} to handle drifting data streams more robustly. However, depending on a single source stream may still degrade performance due to data quality constraints.

To enhance robustness, multisource stream classification exploits complementary information across heterogeneous input streams~\cite{hou2025osasformer}. Early frameworks like Melanie~\cite{du2019multi} employed weighted ensembles to transfer knowledge across sources but were restricted to supervised settings. Subsequent approaches introduced unsupervised node-weighting~\cite{renchunzi2022automatic} and multi-objective optimization (MCMO)~\cite{jiao2022reduced} to mitigate covariate shift and identify shared feature subsets. More recently, the focus has shifted toward dynamic correlation modeling. OBAL~\cite{yu2024online} dynamically tracks inter-stream correlations for adaptive online learning, while BFSRL~\cite{yu2024fuzzy} learns fuzzy shared representations across streams to capture latent commonalities under complex drift.

ii) Multistream collaborative prediction exploits complementary information across multiple evolving streams for joint forecasting. Early approaches focus on adaptive fusion: Wang et al.~\cite{wang2024adaptive} propose a selective stacking mechanism that retrains component models only when drift-induced degradation is detected, while Wen et al.~\cite{wen2023onenet} design a dual-branch framework that disentangles temporal dynamics from cross-variable dependencies. CORAL~\cite{xu2025coral} further advances this direction by employing kernel-induced self-representation to model the co-evolving structure of time series, using collective historical patterns as a stable reconstruction basis.
Moving beyond simple fusion, recent work emphasizes explicit modeling of inter-stream correlations as a structural prior for adaptation. Zhang et al.~\cite{zhang2025multistream} use fuzzy logic to quantify the usefulness of non-drifting streams, enabling stable knowledge sharing that mitigates drift in affected streams. To account for correlation drift—i.e., changes in the relationships among streams—Evolutionary Regressor Chains~\cite{zhang2025tracking} introduce a heuristic order-search procedure that dynamically updates the chain structure during online learning. Elevating this structural perspective, Zhou et al.~\cite{zhou2023multi} represent streams as nodes within a dynamic graph, enabling subgraph-level prediction and localized adaptation. Their subsequent work~\cite{zhou2025continuous} extends this idea by learning the graph topology itself in an online manner, allowing structural dependencies to evolve continuously with the data.

Furthermore, a key challenge in real-world multistream scenarios is heterogeneity, where streams may differ in feature spaces, temporal resolution, or semantic spaces. To address this, Wang et al.~\cite{wang2025Adaptive} propose the AIF-CD framework, which dynamically aligns heterogeneous representations during drift. Complementarily, Yu et al.~\cite{yu2025drift} introduce the modular CAMEL framework, assigning each stream a specialized expert with its own feature extractor while enabling coordinated cross-expert knowledge exchange. This modularization balances stream-specific specialization with cross-stream collaboration, leading to more robust adaptation under complex, heterogeneous environments.

\subsection{Rhythmic Drift}
% Predictable Time Drift
% 通用解决框架及代表方法
Unlike arbitrary drift, which occurs in an unpredictable or stochastic manner, rhythmic drift describes concept evolution that follows a structured, periodic, or smoothly varying temporal pattern~\cite{nasery2021training}. Such drift is particularly common in real-world time-series and streaming scenarios where data generation is influenced by intrinsic temporal regularities, such as daily cycles, seasonal variations, or user behavior rhythms. For example, electricity consumption peaks during certain hours, stock market volatility exhibits weekly trends, and environmental data often follow seasonal changes. In these contexts, the underlying joint distribution $p_t(X, y)$ does not change abruptly or randomly, but evolves according to a predictable rule, satisfying $p_{t+T}(X, y) \approx p_t(X, y)$ for some cycle length $T$~\cite{yu2025learning}. Recognizing and exploiting such regularity enables models not only to respond to drift but also to anticipate it, forming the basis for proactive adaptation and long-term generalization. A substantial portion of real-world non-stationarity is driven by recurring or trend-consistent patterns, which degrade predictive performance if treated solely through reactive retraining or window-based adaptation. This observation motivates a paradigm shift from detecting drift events to modeling their temporal evolution, positioning rhythmic drift as a first-class modeling target rather than a by-product of distributional change.

Rhythmic drift introduces a distinct challenge and opportunity for adaptive learning. Traditional drift detection and adaptation methods treat distributional change as unexpected, often relying on reactive mechanisms such as error monitoring, hypothesis testing, or window resets. While effective for arbitrary drift, these strategies underutilize temporal regularities that could guide model anticipation. By contrast, rhythmic drift can be understood as a temporally governed process, where the evolution of $p_t(y|X)$ follows a smooth or cyclic trajectory. This realization bridges the field of concept drift with temporal domain generalization (TDG) and time-series forecasting , motivating algorithms that explicitly encode time-dependent structure to achieve forward-looking generalization.

\subsubsection{Temporal Domain Generalization.}
Recent advances in TDG aim to model such temporally evolving domains by learning representations that remain stable across time while preserving sensitivity to temporal dynamics. For instance,  Concept Drift Simulation for Temporal Domain Generalization (CODA)~\cite{chang2023coda} simulates future domain distributions through learned drift trajectories, enabling models to anticipate unseen temporal conditions. Similarly, Drift-Aware Dynamic Neural Network (DRAIN)~\cite{nasery2021training} introduces a dynamic graph-based architecture that models the continuous evolution of task-specific parameters over time, effectively capturing the rhythm of gradual drift. Another representative work, Concept Drift Representation Learning for Co-evolving Time Series (CORAL)~\cite{xu2025drift2matrix}, proposes a representation learning framework that explicitly disentangles stable and drifting factors in multivariate time-series, allowing the model to track and reuse recurring patterns. FreKoo~\cite{yu2025learning} introduces a novel frequency-aware perspective, modeling long-term periodicity and achieving robustness against domain-specific uncertainties. These works collectively demonstrate a growing shift from reactive adaptation to proactive generalization in temporally structured environments.

Furthermore, recognizing the continuous nature of time, Continuous TDG (CTDG) methods have been proposed, often treating time as a continuous index for adaptation~\cite{cai2024continuous, wang2020continuously}. EvoS~\cite{xie2024evolving} proposes a multi-scale attention module (MSAM) to model evolving feature distribution patterns across sequential domains, dynamically standardizing features using predicted statistics to mitigate distribution shifts while employing adversarial training to maintain a shared feature space and prevent catastrophic forgetting. Koodos~\cite{cai2024continuous} models data and model dynamics as a continuous-time system via Koopman operator theory. It integrates prior knowledge and multi-objective optimization to synchronize model evolution with data drift. ~\cite{qin2023evolving} propose MMD-LSAE, which leverages an autoencoder architecture to model the evolving patterns among domains in the latent space. Static-DYNamic Causal Representation Learning (SYNC)~\cite{he2025learning}, an approach that effectively learns time-aware causal representations and reconstruct causal mechanisms to address distribution shifts over time. 

\subsubsection{Time-series Forecasting.}
Rhythmic drift is a pervasive phenomenon in non-stationary time-series analysis. Early works on rhythmic drift emphasize learning the dynamics of drift itself rather than merely responding to its consequences~\cite{abdullahi2025systematic}. Learning to Learn the Future~\cite{you2021learning} adopts a meta-learning perspective, where the model learns transition patterns between temporal regimes, enabling anticipation of future drift trajectories. Building on this idea, DDG-DA~\cite{li2022ddg} introduces a generative framework that explicitly synthesizes future data distributions based on historical drift evolution, allowing models to adapt preemptively rather than reactively. In addition, many works address non-stationarity through frequency-aware modeling, decomposition, and adaptive normalization. FEDformer~\cite{zhou2022fedformer} enhances long-term forecasting by decomposing time series into trend and seasonal components and selectively modeling sparse frequency representations. ATFNet~\cite{li2023atfnet} further integrates time-domain and frequency-domain features via adaptive ensembling, dynamically balancing local temporal dependencies and global periodic structure. From a normalization perspective, SAN~\cite{liu2023adaptive} mitigates non-stationarity by performing adaptive normalization over temporal slices rather than the entire sequence, while FAN~\cite{ye2024frequency} extends this idea into the frequency domain by aligning dominant spectral components. Complementarily, Koopman-based predictors~\cite{liu2023koopa} model non-stationary dynamics through latent linear operators, providing a principled mechanism to separate time-invariant structure from evolving temporal dynamics. These methods implicitly exploit rhythmic structure to stabilize long-horizon forecasting under smooth or recurring drift.

More recent methods focus on efficient online adaptation and structural sensitivity to rhythmic changes. OneNet~\cite{wen2023onenet} proposes an online ensembling strategy that continuously reweights subnetworks in response to evolving temporal patterns, achieving robustness without storing extensive historical models. KAN4Drift~\cite{xu2024kan4drift} further explores the use of Kolmogorov–Arnold Networks to detect and track fine-grained functional changes in time series, demonstrating heightened sensitivity to smooth yet persistent drift. Complementarily, Proactive Model Adaptation for Online Time-Series Forecasting~\cite{zhao2025proactive} formalizes a forecast-before-failure paradigm, where latent drift signals are predicted ahead of performance degradation and used to guide anticipatory updates. Collectively, these works indicate a clear trajectory toward proactive, rhythm-aware adaptation, where models synchronize with predictable temporal evolution rather than repeatedly resetting after drift is detected.

\section{Data Stream Drift}

\subsection{Preliminary}

\noindent\textbf{Definition.}
Data stream drift characterizes the evolution of the underlying data-generating process, manifesting as variations in the statistical properties of the input features and their relationship with the target concepts.
Formally, let $\mathcal{X}$ denote the input feature space and $\mathcal{Y}$ the label space.
For a learning task at time step $t$, the data is drawn from a joint probability distribution $P_t(X, Y)$, where $X \in \mathcal{X}$ and $Y \in \mathcal{Y}$.
Data stream drift is said to occur between time $t$ and $t+1$ if the joint distribution changes, i.e., $P_t(X, Y) \neq P_{t+1}(X, Y)$.
According to the Bayesian decision theory, this joint distribution can be decomposed as $P_t(X, Y) = P_t(X) \cdot P_t(Y|X)$.
Consequently, a shift in $P_t(X, Y)$ can originate from changes in the marginal distribution of features $P_t(X)$, the posterior distribution of concepts $P_t(Y|X)$, or a combination of both.
Distinguishing the source of this variation is critical, as it dictates whether the learning system requires feature alignment or decision boundary adaptation.

\noindent
Taxonomy.
While classical drift literature often broadly classifies these changes as ``virtual'' or ``real'' drift, such binary distinctions may oversimplify the complex dynamics observed in high-dimensional and heterogeneous environments.
Motivated by the geometric and semantic nature of these shifts (as illustrated in Fig.~\ref{fig:cases}), we propose a structured taxonomy that categorizes data stream drift into two orthogonal dimensions: representation drift and semantic drift.

Representation Drift arises from shifts in the marginal feature distribution $P(X)$ while the fundamental decision concept $P(Y|X)$ remains relatively stable.
This phenomenon, often termed covariate shift or virtual drift, typically reflects changes in the observation environment—such as sensor noise, domain shifts (e.g., synthetic-to-real), or the emergence of long-tailed distributions.
In these scenarios, the ``appearance'' of the data changes, causing a mismatch between the source and target domains, yet the underlying meaning of the classes remains consistent.

Semantic Drift, in contrast, refers to changes in the posterior distribution $P(Y|X)$, signifying a fundamental shift in the decision boundary itself.
This corresponds to real drift, where the relationship between features and labels evolves—due to redefining class criteria, label noise, or evolving task objectives.
Under semantic drift, the ``meaning'' of the data changes; a feature vector that was previously classified as positive may become negative in the new context.
By separating data stream drift into these two components, we provide a unified lens to analyze how autonomous learning systems must concurrently align their feature representations and update their semantic understanding to maintain robustness in open environments.

\begin{figure}[htpb]
    \centering
    \includegraphics[width=0.45\textwidth]{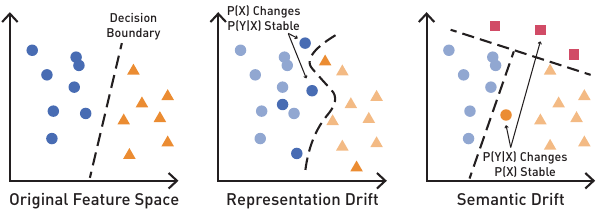}
    \caption{Geometric illustration of data stream drift. (a) Original feature space. (b) Representation drift arises from shifts in the feature marginal $P(X)$, such as domain shift and long-tailed problems, where data points deviate from the source manifold, requiring distributional alignment. (c) Semantic drift arises from shifts in the posterior $P(Y|X)$, where the decision boundary itself evolves, necessitating dynamic decision remapping.}
    \label{fig:cases}
\end{figure}

\subsection{Representation Drift}

\noindent
Representation drift refers to the deterioration in model performance caused by shifts in the marginal feature distribution $P(X)$, even when the conditional relationship $P(Y|X)$ remains stable.
This phenomenon, often termed covariate shift, is prevalent in scenarios such as long-tailed learning, transfer learning, and open-world domain shifts, where the geometric structure of the input data deviates from the training distribution.

To systematically address this challenge, we categorize existing methodologies into three complementary paradigms:
Manifold Reconstruction, which focuses on enriching and repairing the underlying data manifold through synthesis and retrieval;
Decoupling Representation, which aims to isolate robust feature learning from decision biases;
and Distribution Adaptation, which explicitly minimizes the discrepancy between source and target domains through geometric or statistical alignment.

\subsubsection{Manifold Reconstruction}

To mitigate representation drift, manifold reconstruction focuses on enriching or regularizing the underlying data manifold through augmentation, synthesis, or retrieval mechanisms. A primary direction involves generative augmentation to bolster minority classes. Chen et al.~\cite{chenDelvingTrajectoryLongtail2024} employed viewpoint-aware data augmentation to reconstruct feature manifolds for tail classes in multi-target tracking. In the generative domain, Ho et al.~\cite{hoLongTailedAnomalyDetection2024} developed a conditional Variational Autoencoder (cVAE) to explicitly synthesize features for rare categories, while Parashar et al.~\cite{parasharNeglectedTailsVisionLanguage2024} leveraged Large Language Models (LLMs) to address severe representation imbalances in Vision-Language Models (VLMs).

Beyond direct generation, recent research emphasizes optimizing the augmentation process itself. Wang et al.~\cite{wangKillTwoBirds2023} critically examined the trade-offs of augmentation strategies under long-tailed distributions, proposing an optimization framework that dynamically adjusts augmentation intensity.
Building upon this, Li et al.~\cite{liSynthesizingMinoritySamples2025} introduced distribution-matching synthesis to ensure generated samples align with the true minority distribution.

Alternatively, retrieval and sampling strategies offer non-generative solutions. Jung et al.~\cite{jungTailedCoreFewShotSampling2025} proposed TailedCore, a few-shot sampling strategy that prioritizes informative minority exemplars. Expanding the scope to open-world data, Sidhu et al.~\cite{sidhuSearchDetectTrainingFree2025} adopted a training-free approach that enriches tail distributions by retrieving relevant images from web-scale databases.

\subsubsection{Decoupling Representation}

Decoupling-based strategies fundamentally aim to isolate the learning of robust feature representations from the biases inherent in decision boundaries. A primary line of research scrutinizes training dynamics to mitigate classifier overfitting. Sun et al. \cite{sunRethinkingClassifierReTraining2024} reveal that the classifier head is prone to overfitting head-class features and propose a label over-smoothing mechanism to recalibrate the decision surface. Building on this insight, distillation has emerged as a potent decoupling tool; Rangwani et al. \cite{rangwaniDeiTLTDistillationStrikes2024} introduce DeiT-LT to address representation drift in Vision Transformers, while Deng et al. \cite{dengEIFAKDExplicitImplicit2026} develop EIFA-KD, which synergizes explicit and implicit feature augmentation with knowledge distillation to enhance generalization.

Parallel efforts focus on restructuring the feature topology through prototype alignment and distribution regularization. To ameliorate representation drift caused by intra-class data imbalance, Wang et al. \cite{wangLongTailClassIncremental2024} construct independent sub-prototype spaces, whereas Guo et al. \cite{guoPrototypeAlignmentDedicated2025} employ prototype alignment with dedicated experts to minimize intra-class dispersion. From a distributional perspective, Zheng et al. \cite{zhengBEMBalancedEntropybased2024} propose the Balanced Entropy-based Mix (BEM) to rebalance feature distributions, ensuring a more uniform coverage of the latent space.

From the perspective of representation geometry, recent works enforce structural constraints to maximize discriminability. Yang et al. \cite{yangTdistributedSphericalFeature2023} introduce a $T$-distributed spherical feature representation to regularize the manifold structure, while Du et al. \cite{duProbabilisticContrastiveLearning2024} leverage probabilistic contrastive learning to maximize inter-class separability. Zhou et al. \cite{zhouContinuousContrastiveLearning2024} further extend this paradigm to semi-supervised settings via continuous contrastive learning. To enhance representation robustness under partial observability, Yang et al. \cite{yangone} propose an occlusion-based contrastive learning framework. By constructing semantic-aware views that compel the model to infer global context from local fragments, this approach fosters representations that remain stable even when visual inputs are partially occluded or corrupted, effectively mitigating representation drift induced by feature loss.

\subsubsection{Distribution Adaptation}

Distribution adaptation mitigates representation drift by explicitly aligning disparate distributions into a shared latent space. A foundational research direction addresses the temporal nature of drift. To capture the evolution of target distributions, He et al.~\cite{heDomainAdaptationTime2023} investigate time-dependent domain adaptation strategies. Complementing this, Saberi et al.~\cite{saberiGradualDomainAdaptation2024} introduce gradual domain adaptation, which effectively transfers knowledge across intermediate stages of drift, thereby bridging the gap between source and evolving target domains.

Beyond temporal aspects, significant efforts have been directed toward complex structural alignment in multi-source scenarios. Feng et al.~\cite{fengOpenCompoundDomain2023} tackle the challenges of open compound domains, while Chen et al.~\cite{chenMultiPromptAlignmentMultiSource2023} leverage multi-prompt learning to facilitate versatile alignment. From a geometric perspective, spectral and subspace methods offer structural stability: Xiao et al.~\cite{xiaoSPAGraphSpectral2023} utilize spectral graph alignment, and Li et al.~\cite{liSubspaceIdentificationMultiSource2023} identify shared subspaces among multiple domains. Building upon these geometric foundations, Chang et al.~\cite{changUnifiedDomainGeneralization2024} propose a unified framework for domain adaptation and generalization. Furthermore, probabilistic approaches provide rigorous theoretical grounding; Wang et al.~\cite{wangFDivergencePrincipledDomain2024} establish a framework based on $f$-divergence minimization, while Du et al.~\cite{duDiffusionBasedProbabilisticUncertainty2023} employ diffusion-based probabilistic modeling to quantify uncertainty amidst distribution shifts.

Adversarial and contrastive learning paradigms have emerged as powerful tools for enforcing feature invariance. Sun et al.~\cite{sunAdversarialAlignmentAnchor2025} advance this field with an anchor-dragging adversarial alignment mechanism. Concomitantly, Liu et al.~\cite{liuBoostingTransferabilityDiscriminability2024} enhance cross-domain performance by jointly optimizing transferability and discriminability. Addressing specific drift nuances, Lu et al.~\cite{luStyleAdaptationUncertainty2024} focus on style-induced drift via uncertainty calibration. In the realm of dynamic visual data, Wei et al.~\cite{weiUnsupervisedVideoDomain2023} adapt temporal representations by simultaneously aligning motion and appearance distributions.

Finally, in distributed environments where privacy is paramount, federated adaptation has garnered significant interest. Shi et al.~\cite{shiCLIPGuidedFederatedLearning2024} incorporate CLIP-guided global representations to enhance federated learning robustness. Similarly, Jia et al.~\cite{jiaDapperFLDomainAdaptive2024} propose Dapper-FL, a method designed to meticulously balance the trade-off between local client adaptation and global model aggregation.

\subsection{Semantic Drift}
\label{sec:semantic-drift}

\noindent
Semantic drift represents a critical frontier in understanding how learning systems maintain meaning under evolving contexts.  
While representation drift focuses on the instability of feature spaces, semantic drift captures how the association between inputs and outputs—the semantics of prediction—shifts over time.  
These shifts may arise from changing annotation rules, evolving task objectives, or dynamic environments that redefine what a label signifies.  
Such evolution undermines the stationarity assumption of $P(Y)$, $P(Y|X)$, or $P(X|Y)$, leading to gradual degradation in both model reliability and interpretability.  

This section surveys methodological advances that explicitly address semantic drift, through four interrelated perspectives: 
i) Invariant Representation Learning focuses on stability. It seeks to immunize models against drift by anchoring representations to causal, structural, or geometric regularities that remain constant even as superficial environmental contexts evolve.
ii) Adaptive Continual Learning emphasizes plasticity. Accepting that invariance has limits, these methods design models that actively co-evolve with the data stream, utilizing mechanisms like test-time adaptation and meta-learning to realign decision boundaries in real-time.
iii) Probabilistic Rectification adopts a diagnostic stance. It treats semantic drift as a stochastic process, leveraging generative modeling, uncertainty estimation, and energy-based frameworks to quantify, interpret, and rectify the degradation of semantic confidence.
iv) Semantics Synchronization addresses the collective dimension. In decentralized or federated settings, these approaches ensure global semantic coherence across distributed agents, preventing the fragmentation of meaning that arises when local clients evolve asynchronously.

\subsubsection{Invariant Representation Learning}

The quintessential challenge in managing semantic drift lies not merely in change detection, but in the preservation of semantic integrity. 
Invariant representation learning endeavors to encode semantics that remain stable despite evolving environments or fluctuating task definitions. 
Departing from reactive adaptation strategies, these methods proactively embed invariance within the feature space, anchoring semantics to causal or structural regularities that persist across domains.

Causality-inspired paradigms have emerged as a cornerstone in this domain. Kim et al.~\cite{kimSufficientInvariantLearning2025} introduce Sufficient Invariant Learning, which rigorously disentangles invariant causal mechanisms from spurious correlations, ensuring semantic stability even when context-dependent associations shift. Complementing this, Li et al.~\cite{liLetInvariantLearning2025} propose neighbor-shift generalization to isolate stable local dependencies. Collectively, these studies exemplify a fundamental solution to semantic drift: rather than rectifying drift post-hoc, they mitigate its root cause by encoding environment-agnostic semantics.

Beyond causality, leveraging intrinsic structural and geometric properties is pivotal for sustaining semantic coherence. In the topological domain, Li et al.~\cite{liGraphStructureExtrapolation2024} extrapolate relational dependencies to maintain consistent node semantics amidst changing graph topologies, while Zou et al.~\cite{zouGeSSBenchmarkingGeometric2024} demonstrate how geometric consistency serves as an anchor for semantics across spatial domains. Temporal invariance is equally critical; Liu et al.~\cite{liuTimeseriesForecastingOutofdistribution2024} and Qin et al.~\cite{qinEvolvingDomainGeneralization2023} advance this by learning time-invariant embeddings that preserve evolving temporal semantics.

Furthermore, recent scholarship addresses the degradation of meaning under label noise and contextual reinterpretation. Yu et al.~\cite{yuTreatNoiseDomain2023} treat label corruption as a domain shift, disentangling true label semantics from noisy supervision. Building on this, Yang et al.~\cite{yangReCDAConceptDrift2024} operationalize semantic alignment through ReCDA, constraining latent representations to remain robust as concepts drift. In summary, invariant representation learning seeks to immunize models against drift by integrating causal, structural, and temporal invariance directly into the representation manifold.

\subsubsection{Adaptive Continual Learning}

While invariant representation learning strives to resist change, adaptive continual learning embraces it. When invariance becomes insufficient to capture drastic distribution shifts, adaptability becomes paramount. These methods fundamentally reconceptualize meaning not as a static property to be preserved, but as a fluid entity that necessitates continuous reinterpretation through dynamic parameter adjustment.

For immediate, real-time semantic correction, Test-Time Adaptation (TTA) has emerged as a lightweight yet potent paradigm. Kim et al.~\cite{kimTestTimeAdaptationInduces2024} demonstrate that entropy minimization during the inference phase can effectively realign decision boundaries to accommodate unseen semantics. Building on this, Yang et al.~\cite{yangDualTestTimeTraining2025} refine the adaptation process via dual test-time training, which integrates instance-level constraints with memory-based regularization. Similarly, Yang et al.~\cite{yangOODDTesttimeOutofDistribution2025} introduce dynamic dictionaries to update semantic prototypes online. 
Collectively, these approaches embody an immediate responsiveness, crucial for autonomous learning systems operating in rapidly fluctuating environments.

In contrast, addressing long-term semantic evolution requires persistent and structural adaptation mechanisms. Zhao et al.~\cite{zhaoProactiveModelAdaptation2025} advance a proactive framework that anticipates drift, updating the model preemptively before performance degradation occurs. Moreover, Liu et al.~\cite{liuUDDATCUnsupervisedRealTime2025} facilitate unsupervised adaptation by coupling drift detection with online learning updates, while Zhang et al.~\cite{zhangAdaO2BAdaptiveOnline2025} bridge the gap between online and batch processing to ensure smooth temporal recalibration. These architectures signify a paradigm shift toward plastic models—systems designed to co-evolve with the data stream rather than rigidly withstanding it.

Finally, to enhance adaptation velocity, meta-learning and few-shot paradigms have gained prominence. Aguiar and Cano~\cite{aguiarEnhancingConceptDrift2023} leverage meta-learned priors to accelerate the detection of drift, whereas Rahman et al.~\cite{rahmanDecouplingClinicalClassAgnostic2026} demonstrate the capability for few-shot semantic reinterpretation using minimal examples. These methods emphasize the concept of learning to adapt, abstracting the adaptation process itself into a transferable skill that enables rapid recovery from semantic shifts.

\subsubsection{Probabilistic Rectification}

Probabilistic rectification adopts a diagnostic stance toward semantic drift.
Unlike methods that merely constrain feature spaces or adapt parameters reactively, these approaches seek to elucidate drift by modeling the stochastic fluctuations of semantics themselves. By rigorously capturing uncertainty and inferring latent ground truths, they offer interpretability—unveiling not only that a shift has occurred, but also quantifying the mechanics of how and why semantics have evolved.

Generative modeling serves as a robust mechanism for reconstructing degraded semantics. Na et al.~\cite{naLabelNoiseRobustDiffusion2023} leverage diffusion models to recover true labels from noisy annotations, treating semantic corruption as a reversible generative process. Extending this to structural domains, Shen et al.~\cite{shenOptimizingOODDetection2024} demonstrate that generative consistency can preserve semantic integrity in molecular data, even amidst significant structural perturbations.

Complementing reconstruction, uncertainty estimation provides a metric for semantic instability. Mehrtens et al.~\cite{mehrtensBenchmarkingCommonUncertainty2023} benchmark uncertainty under label noise, establishing prediction entropy as a critical early warning signal for semantic misalignment. Furthermore, Neo et al.~\cite{neoMaxEntLossConstrained2024} impose entropy regularization to stabilize semantic confidence, preventing the model from becoming overconfident in drifting environments. These probabilistic measures constitute the diagnostic stratum upon which adaptive and invariant strategies can be effectively deployed.

Advanced frameworks further bridge these paradigms through energy-based and trajectory modeling. Tian et al.~\cite{tianModelingDistributionalUncertainty2023} utilize energy landscapes to delineate semantic boundaries, while Wang et al.~\cite{wangEmbeddingTrajectoryOutofDistribution2024} explicitly model the temporal evolution of semantic trajectories. Additionally, Farid et al.~\cite{faridTaskDrivenDetectionDistribution2025} integrate task-driven uncertainty estimation to track meaning shifts across dynamic distributions.

In essence, probabilistic rectification transforms semantic drift from an unpredictable failure mode into a tractable stochastic process. By modeling the motion of meaning as a probabilistic evolution, these frameworks provide the necessary calibration and insight to complement invariant and adaptive solutions.

\subsubsection{Semantics Synchronization}

In decentralized learning environments, semantic drift becomes a collective phenomenon. Different clients may evolve their local semantics independently, leading to fragmented or inconsistent global models. Federated and collaborative adaptation thus focuses on aligning semantic meaning across distributed learners without requiring direct data sharing.

Prototype and aggregation based designs form the foundation of this paradigm. Mawuli et al.~\cite{mawuliFedStreamPrototypeBasedFederated2023} introduce FedStream, which maintains semantic prototypes across clients to synchronize evolving local representations. This strategy prevent semantic fragmentation across asynchronously drifting participants.

A complementary direction emphasizes cross-client invariance and relational generalization. Liao et al.~\cite{liaoFOOGDFederatedCollaboration2024} develop FOOGD, promoting invariant representation sharing for federated OOD generalization, and Zhou et al.~\cite{zhouFedGOGFederatedGraph2025} extend it to graph-structured data through cross-client semantic alignment. Both frameworks demonstrate that federated learning can sustain consistent semantics by aligning relational meaning rather than raw features.

Recent advances integrate meta-learning and multi-agent coordination to adapt semantics collaboratively. Zhao et al.~\cite{zhaoFedFMFederatedFewshot2025} introduce FedFM, enabling few-shot adaptation of semantics across distributed nodes. Together, these approaches show that semantic drift in decentralized settings is not merely a modeling challenge, but a communication and coordination problem.

In summary, federated and collaborative drift adaptation extends the fight against semantic drift from the individual model to the collective system.  
By coupling local adaptability with global coherence, these methods provide the organizational infrastructure for autonomous learning systems that must continually negotiate and synchronize meaning across diverse, evolving contexts.

\section{Model Stream Drift}

\subsection{Preliminary}

\noindent\textbf{Definition.}
Model stream drift refers to the internal evolution, divergence, or instability of the learning system's hypothesis parameters and structural configurations over time or across agents.
Unlike data stream drift, which originates from external changes in the input distribution $P(X, Y)$, model stream drift is intrinsic to the learner itself.
Formally, let a learning model be defined as a function $f_{\theta}: \mathcal{X} \rightarrow \mathcal{Y}$, parameterized by $\theta \in \Theta$.
Model stream drift occurs when the parameter state shifts significantly from $\theta_t$ to $\theta_{t+\Delta}$ (temporal evolution), or when parameters diverge across distinct model instances $\theta^{(i)} \neq \theta^{(j)}$ (spatial divergence), such that the decision boundary is altered not solely by data, but by the optimization trajectory, forgetting mechanisms, or feedback loops.
This internal dynamism challenges the conventional view of a static ``optimal'' model, necessitating mechanisms that govern the stability, plasticity, and consensus of $\theta$.

\noindent
Taxonomy.
To systematically analyze these internal dynamics, we categorize model stream drift into three distinct dimensions based on the nature of the parameter evolution and the learning context: sequence drift, heterogeneity drift, and policy drift.

Sequence Drift captures the temporal evolution of a single model's parameters $\theta_{\tau} \to \theta_{\tau+1}$ as it learns continuously from a stream of tasks.
The core challenge here is the stability-plasticity dilemma: as the model updates $\theta$ to acquire new knowledge (plasticity), it risks overwriting the optimal configuration for previous concepts, leading to catastrophic forgetting.
This drift describes the tension between maintaining historical knowledge and adapting to current inputs.

Heterogeneity Drift characterizes the spatial divergence of parameters across multiple distributed model instances $\{\theta^{(1)}, \dots, \theta^{(K)}\}$.
Common in federated and decentralized learning, this drift arises when local models evolve in different directions due to non-IID local data or differing architectures.
Here, the goal is not just individual adaptation, but managing the aggregation and synchronization of diverse parameter states to prevent system fragmentation.

Policy Drift describes the endogenous instability in decision-making policies, particularly in reinforcement learning and feedback-loop systems.
It occurs when the current target policy $\pi_\theta$ deviates excessively from the behavior policy $\pi_{old}$ used to collect data, or when the model's actions fundamentally alter the environment's state distribution.
Unlike sequence drift which is driven by new external data, policy drift is often self-induced, where the agent's own updates lead to distributional mismatch and optimization instability.

\subsection{Sequence Drift}

\begin{figure}[htbp]
    \centering
    \begin{subfigure}[t]{0.48\textwidth}
        \centering
        \includegraphics[width=0.98\textwidth]{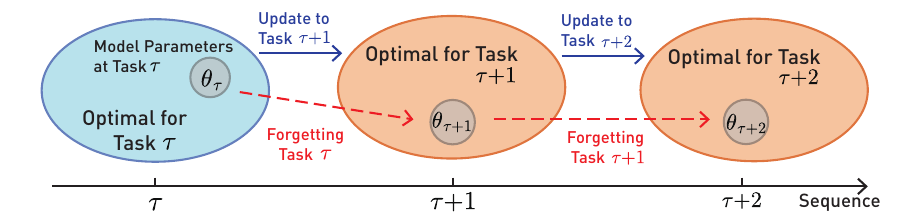}
        \caption{Sequence Drift}
        \label{fig:seq-adap}
    \end{subfigure}
    \begin{subfigure}[t]{0.48\textwidth}
        \centering
        \includegraphics[width=0.98\textwidth]{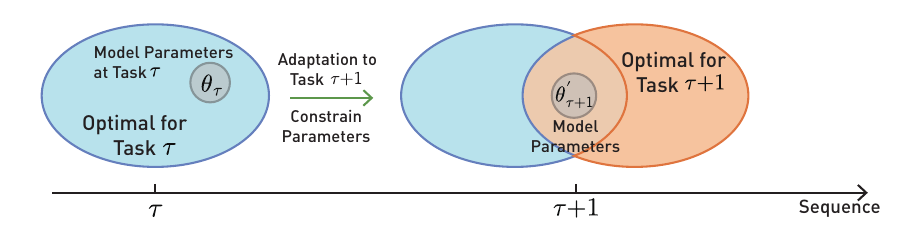}
        \caption{Stability-Plasticity Balance}
        \label{fig:seq-adap}
    \end{subfigure}
    \caption{The Stability-Plasticity Balance in Sequence Drift. The parameter evolution $\theta_{\tau} \to \theta_{\tau+1}$ faces a conflict between Adaptation (learning new tasks) and Forgetting (losing old patterns). A central Meta-Stability Module mediates this tension by regulating updates via Parameter Space Regularization, Structural Space Expansion, or Optimization Space Meta-Learning, ensuring continuous evolution without catastrophic failure.}
    \label{fig:analysis}
\end{figure}

As models evolve within continuous or temporally correlated data streams, their internal representations and decision boundaries undergo gradual transformation, a phenomenon broadly referred to as sequence drift. Unlike data drift, which arises from distributional changes in inputs, sequence drift reflects the dynamic evolution of the model itself—how its parameters, features, and reasoning trajectories shift over time as it learns from non-stationary experiences. The central challenge lies in achieving a stable equilibrium between adaptation to new information and retention of accumulated knowledge. Excessive plasticity accelerates forgetting, while over-consolidation hinders responsiveness. Recent studies therefore reframe model drift as a continuous regulation problem, where learning stability emerges not from static optimization but from dynamic interaction between memory preservation and adaptive flexibility. From this perspective, sequence drift can be understood through three interdependent methodological pathways: parameter regularization, dynamic architectures, and optimization meta-learning.

\begin{table}[htbp]
\centering
\begin{tabular}{@{}llp{100px}@{}}
\toprule
Methods   Types                                                       & Intervention   Object    & Mechanism                                                                            \\ \midrule
\begin{tabular}[c]{@{}l@{}}Parameter\\ Regularization\end{tabular}    & Weights ($\theta$)       & Constraining: Limit the gradient   $\Delta\theta$ to protect important weights.      \\
\begin{tabular}[c]{@{}l@{}}Structural\\ Expansion\end{tabular}        & Capacity ($\mathcal{H}$) & Expanding: Add model parameters   $\theta_{new}$ to physically isolate interference. \\
\begin{tabular}[c]{@{}l@{}}Optimization \\ Meta-Learning\end{tabular} & Trajectory ($\nabla L$)  & Steering: Find the optimal update path for   Pareto.                                 \\ \bottomrule
\end{tabular}
\caption{A taxonomy of intervention mechanisms for Sequence Drift. The strategies are categorized by their intervention objects, Weights ($\theta$), Capacity ($\mathcal{H}$), and Trajectory ($\nabla L$), illustrating how they respectively constrain, expand, or steer the model's evolution to balance stability and plasticity.}
\label{tab:my-table}
\end{table}

\subsubsection{Parameter Regularization}  

In the context of sequence drift, parameter regularization aims to stabilize representational consistency as the model evolves over temporally correlated data streams. Rather than reacting to distributional fluctuations at the data level, consolidation-based approaches constrain the internal transformation of features and decision boundaries to mitigate representational drift within the model itself. Early efforts introduce hierarchical or spatially organized mappings that preserve structural coherence when sequential concepts evolve~\cite{amadorcoelhoConceptDriftDetection2023}. Unsupervised autoencoding strategies further enable latent-space self-alignment, allowing the model to “unlearn” outdated representations and reinforce stable manifolds across time~\cite{arteltUnsupervisedUnlearningConcept2023}. More recent works incorporate deep representation awareness into drift detection and adaptation, leveraging neural embeddings as anchors for stability. In class-incremental learning, several studies explicitly disentangle feature and classifier updates to avoid entangled drift propagation. Dynamic feature-space self-organization and calibration~\cite{heDYSONDynamicFeature2024} jointly reduce catastrophic forgetting and maintain intra-class compactness. Complementary to these, covariance adaptation and relaxed distillation enforce consistent local geometry to resist temporal bias accumulation.
Other consolidation paths adopt prototype-based memory as semantic anchors to maintain class topology and prevent manifold collapse~\cite{sunClassIncrementalLearning2023}. To enhance resilience under sequential updates, model-surgery frameworks reconstruct pre-trained representations with minimal parameter intervention~\cite{sunMOSModelSurgery2025}, while feature translation or exemplar-free paradigms ensure structural invariance without explicit rehearsal~\cite{petitFeTrILFeatureTranslation2023}. 
% In few-shot or graph-incremental scenarios, consolidation is achieved through prototype transfer and inductive structure embedding that preserve long-term relational consistency \cite{liInductiveGraphFewshot2025}. 
Collectively, these methods establish the foundation of model-level drift regulation, where representational stability—not merely accuracy—is treated as the primary objective. By consolidating internal knowledge states, the model develops a temporal equilibrium that enables robust, memory-preserving adaptation across evolving sequential environments.

\subsubsection{Dynamic Architectures}

While consolidation stabilizes representational anchors, dynamic architectures focuses on the model’s ability to flexibly integrate novel information over temporally evolving sequences without losing coherence with prior knowledge. Under continuous data streams, the model must selectively adjust its structure and parameters to accommodate new distributions while preserving learned invariants—a process that defines the essence of sequential model drift. Structural adaptation has emerged as a core strategy, where task-specific or modular components are dynamically integrated to absorb emerging concepts while isolating interference with prior tasks~\cite{liDynamicIntegrationTaskSpecific2025a, petitFeTrILFeatureTranslation2023, sunMOSModelSurgery2025, heDYSONDynamicFeature2024, chenDynamicResidualClassifier2023}. Parameter-level modulation introduces regularization or constrained updates to limit the amplitude of drift in weight space, maintaining a balance between forward plasticity and backward stability~\cite{sunAntiforgettingIncrementalLearning2024, liConceptDriftAdaptation2024a}. Beyond global adjustment, context-sensitive or few-shot mechanisms extend model flexibility to low-data or rapidly shifting regimes, where inductive biases are leveraged to accelerate new concept assimilation \cite{chaiMalFSCILFewShotClassIncremental2025,dengCentroidGuidedDomainIncremental2024}. In dynamic environments, temporal consistency and bounded updates enable reliable adaptation to nonstationary sensory inputs or drifting process conditions, as reflected in industrial, cyber, and IoT applications \cite{zhouIndustrialFaultDiagnosis2025, pengUnsupervisedAdaptiveFleet2024a}.
Collectively, these studies converge on the notion that model drift resilience does not emerge from rigidity, but from controlled plasticity—the capacity to locally deform decision boundaries in proportion to environmental dynamics. Dynamic architectures thus transforms sequential learning from reactive adjustment into a continuous, structurally adaptive process that internalizes change as part of the model’s evolving equilibrium.

\subsubsection{Optimization Meta-Learning}

Beyond consolidation and dynamic architectures, optimization meta-Learning represents a higher-order paradigm in sequential drift research, where the model acquires the ability to regulate its own stability–plasticity balance over long-term evolution. Instead of passively correcting for drift, meta-stable systems internalize the dynamics of change, learning how to learn under temporally correlated uncertainty. At the core of this direction lies meta-learning and transfer-based regulation, which enable models to infer adaptive update rules from prior drift experiences, constructing a meta-space that captures cross-task regularities and parameter trajectories~\cite{yuGeneralizedIncrementalLearning2025a, yangAdaptingMultimodalLarge2024b}. Meanwhile, adaptive ensemble and cooperative frameworks extend stability through population-level coordination, dynamically reweighting or synchronizing sub-models according to real-time drift signals \cite{yuOnlineBoostingAdaptive2024a, zhangOneNetEnhancingTime2023a, liConceptDriftAdaptation2024a}. Such multi-agent or federated mechanisms embody collective meta-stability, where local adaptations are globally regularized to maintain systemic equilibrium across asynchronous drift. Complementary to this, drift-aware optimization frameworks directly encode temporal uncertainty into the learning process, constructing self-referential objectives that adaptively modulate update magnitudes and directions~\cite{yuDetectingGroupConcept2023c, amadorcoelhoConceptDriftDetection2023}. Together, these approaches suggest that model drift management can transcend episodic adaptation toward a continuous, self-evolving process: the model not only perceives and responds to drift but progressively abstracts the principles of change itself. Optimization meta-learning thus closes the loop of sequential adaptation, transforming the model from a reactive learner into an autonomous dynamical system that sustains long-term stability through recursive self-regulation.

In summary, sequence drift manifests as the temporal evolution of model parameters and representations under continuous learning. The three pathways, parameter regularization, dynamic architectures, and optimization meta-learning, outline a progression from passive stability enforcement to proactive self-regulation. Consolidation preserves internal coherence, adaptation enables controlled deformation, and meta-stability encapsulates self-evolving equilibrium. Together they form a unified theoretical spectrum of model drift dynamics, laying the conceptual foundation for subsequent discussions on heterogeneity and policy drift in autonomous learning systems.

\subsection{Heterogeneity Drift}

Heterogeneity Drift refers to the phenomenon where the representations, parameters, or optimization trajectories of multiple model instances diverge over time, even when they are trained toward similar objectives. Unlike data drift—which manifests as changes in the data distribution—heterogeneity drift occurs within or across models, reflecting discrepancies in architectures, learning dynamics, or contextual adaptation. This form of drift is particularly critical for autonomous and distributed systems, where the coordination of multiple heterogeneous learners determines the overall system stability and generalization.

Early studies explored this problem within decentralized and federated environments, where model heterogeneity arises from non-IID data and asynchronous optimization. Yu et al. proposed a parallel–ensemble approach to enhance global generalization under heterogeneous client distributions \cite{yuFederatedLearningAlgorithm2023}, while Wang and Zhang introduced the FedSiM mechanism, which quantifies inter-client similarity through a stimulus–response perspective \cite{wangFedSiMSimilarityMetric2023}. Lian et al. developed GOFL, a gradient optimization-based framework that normalizes client updates to stabilize aggregation in heterogeneous IoT networks \cite{lianGOFLAccurateEfficient2024}.

Recent efforts extend beyond static aggregation toward adaptive heterogeneity management. Panchal et al. introduced Flash, an adaptive framework that detects and compensates for local concept drift during federated updates \cite{panchalFlashConceptDrift2023b}. Ganguly and Aggarwal proposed online adaptation mechanisms that detect non-stationarity and reconfigure model updates dynamically \cite{gangulyOnlineFederatedLearning2024}. Complementary approaches focused on the client side: Saile et al. designed a client-level drift adaptation scheme \cite{saileClientSideAdaptationConcept2024}, and Thomas et al. formulated a selection-based strategy to maintain robustness against heterogeneous drift across participants \cite{thomasAdaptionSelectionClient2025a}.

Beyond client-driven divergence, cross-model heterogeneity also arises in continuous and domain-incremental learning. Babendererde et al. introduced a federated-continual dynamic segmentation strategy guided by Barlow continuity to handle continual model evolution in histopathology \cite{babendererdeFederatedContinualDynamicSegmentation2025}, while Kumari et al. developed privacy-aware domain-incremental learning for digital pathology to accommodate evolving domain shifts \cite{kumariContinualDomainIncremental2024}. Kalb and Beyerer \cite{kalbPrinciplesForgettingDomainIncremental2023} analyze forgetting mechanisms under domain-incremental conditions, showing that structural divergence may accumulate even with shared objectives. Zhang et al. \cite{zhangSpatialTemporalFederated2025} extend lifelong learning to spatial–temporal settings, emphasizing synchronization among distributed edge models.

From the perspective of fairness and medical robustness\cite{yang2024segmentation,yang2022local,young2026xrayclaw}, heterogeneity drift has been shown to interact with population bias and covariate imbalance. Zhu et al. \cite{zhuFedWeightMitigatingCovariate2025} proposed FedWeight, which reweights patient contributions to correct for covariate shift in electronic health records. Chen et al. \cite{chenAlgorithmicFairnessArtificial2023} highlighted how algorithmic fairness constraints can inadvertently amplify model heterogeneity.

Overall, heterogeneity drift represents an intrinsic challenge of modern model ecosystems, spanning from federated to centralized and automated settings. As models evolve, specialize, and interact within dynamic environments, their internal representations and learning dynamics may diverge despite aligned global objectives. Emerging research trends suggest that future solutions will require self-regulating architectures capable of detecting, reconciling, and adapting to structural and behavioral divergence—paving the way for resilient, autonomous learning systems.

\subsection{Policy  Drift}

Policy drift refers to the internal instability that arises when a learning system’s decision policy evolves inconsistently with the data or value functions that guided its previous training. Unlike external distributional shifts or inter-model parameter inconsistencies, policy drift originates endogenously from the agent’s own update dynamics. It is particularly prevalent in off-policy and approximate reinforcement learning algorithms—such as PPO or actor–critic frameworks—where the newly updated policy diverges from the behavior policy used to collect training data, resulting in biased gradients, unstable convergence, or performance collapse. This phenomenon encapsulates the feedback-driven fragility of autonomous learning: as the model continually acts upon and reshapes its environment, even small deviations in policy distribution can amplify across iterations, leading to long-term instability if not properly regulated. Understanding and controlling policy drift has thus become essential for building scalable, self-stabilizing autonomous learning systems.

\subsubsection{Policy Constraints}  
A primary strategy to mitigate policy drift is explicitly constraining the magnitude of policy updates. For instance, Ajay et al.~\cite{ajayDistributionallyAdaptiveMeta2022} introduced a distributionally adaptive meta-reinforcement learning framework that regularizes policy updates with respect to the task distribution. In the actor–critic domain, DriftShield~\cite{caoDriftShieldAutonomousFraud2025} stabilizes learning by dynamically reweighting features and constraining the critic’s response to abrupt actor updates. Similar constraints have proven effective in industrial applications, such as limiting policy step sizes in chemical polishing~\cite{yuRuntoRunControlChemical2020}, utilizing experience replay for temporal smoothness in EV charging~\cite{poddubnyyOnlineEVCharging2023}, and bounding policy changes in edge-cloud fault prediction~\cite{shayestehAutomatedConceptDrift2022a}. Collectively, bounding the update magnitude through KL constraints, feature reweighting, or temporal regularization serves as an essential safeguard against internal policy-induced drift.

\subsubsection{Off-Policy Rectification}

Off-policy rectification addresses the critical discrepancy between the data distribution induced by the historical behavior policy and that of the evolving target policy.
This misalignment is rigorously formalized by Perdomo et al.~\cite{perdomoPerformativePrediction2020} through the lens of performative prediction, which elucidates how model decisions endogenously influence future data generation. Consequently, the underlying distribution requires periodic re-alignment to prevent the accumulation of feedback-driven drift.
Building upon this theoretical foundation, Zhao et al.~\cite{zhaoPerformativeTimeSeriesForecasting2025} extrapolate the framework to time-series forecasting, proposing distributional adjustment mechanisms to mitigate instability caused by prediction-dependent future trajectories.

In the realm of offline reinforcement learning, rectification focuses on mitigating bias from static datasets.
Schweighofer et al.~\cite{schweighoferDatasetPerspectiveOffline2022} scrutinize policy divergence, positing that the stability of off-policy updates necessitates the selective reweighting or filtering of transitions drawn from outdated behavior policies.
Complementing this, Ajay et al.~\cite{ajayDistributionallyAdaptiveMeta2022} introduce distributionally regularized meta-updates that orchestrate the joint adaptation of the policy and its sampling distribution, effectively minimizing off-policy bias.

Furthermore, recent works extend these principles to complex, non-stationary environments.
Wu and Ismail~\cite{wuGeneralizedRISTile2024} develop a generalized framework for wireless communications that dynamically rectifies policy decisions against evolving channel characteristics.
Similarly, Zhang et al.~\cite{zhangOneNetEnhancingTime2023} employ online ensemble and federated strategies to continuously synchronize model distributions with shifting environments.
Collectively, these studies underscore that policy stability is intrinsically linked to the rectification of distribution shifts, establishing off-policy rectification as a pivotal mechanism for counteracting policy-induced drift.

\subsubsection{Dynamic Regulation}

Dynamic regulation represents a paradigm shift from reactive correction to systemic self-governance.
Instead of passively responding to error accumulation, this approach endows learning systems with higher-order feedback mechanisms to explicitly model and anticipate policy drift.
Ajay et al.~\cite{ajayDistributionallyAdaptiveMeta2022} demonstrate that distributionally adaptive meta-reinforcement learning can facilitate self-regulated policy updates, effectively dampening the instability inherent in non-stationary adaptation.
Advancing this principle, Stein and Tomforde~\cite{steinReflectiveLearningClassifier2021} introduce reflective learning classifier systems that monitor internal evolutionary dynamics, autonomously modulating update rates upon detecting drift signals—a process akin to intrinsic homeostasis.

Architectural integration of these regulatory loops has further matured.
Manias et al.~\cite{maniasModelDriftDynamic2023} extend adaptive mechanisms to dynamic networks, employing self-regulating modules to counteract long-term drift through continual parameter recalibration.
Similarly, the actor–critic framework by Cao et al.~\cite{caoDriftShieldAutonomousFraud2025} and the adaptive reinforcement scheme by Shayesteh et al.~\cite{shayestehAutomatedConceptDrift2022a} embed reflective subloops that scrutinize learning stability.
By blending meta-adaptation with constrained optimization, these methods transform policy optimization into a proactive regulation mechanism, ensuring sustained autonomy before instability accumulates.

Beyond mere stabilization, recent scholarship introduces a normative dimension to regulation.
Yang et al.~\cite{yang2025walking} challenge the monolithic view of drift as strictly detrimental, particularly in the non-stationary custom-tuning of multi-modal large models.
By leveraging counterfactual preference optimization, they propose a nuanced framework to disentangle beneficial drifts—which enhance alignment with evolving tasks—from detrimental drifts that erode reasoning consistency.
This marks a critical evolution in dynamic regulation: moving from suppressing all deviations to selectively cultivating those that serve the system's evolving objectives.

In summary, the mitigation of policy drift requires a balanced integration of stability constraints, distributional alignment, and self-adaptive regulation. Policy constraints provides a direct safeguard against abrupt internal divergence by bounding the policy’s step size during optimization. Data and distribution correction mechanisms further ensure consistency between the behavior and target policies, preserving the validity of off-policy learning. Meanwhile, meta- and reflective adaptation introduces a higher-order layer of self-regulation, allowing models to anticipate and compensate for drift before instability propagates. Together, these strategies highlight a paradigm shift in policy learning—from static optimization toward dynamically self-consistent adaptation—laying the foundation for stable autonomous learning systems to support decision-making in non-stationary environments.

\section{Applications}

The theoretical advancements in drift learning have catalyzed the deployment of autonomous learning systems across a variety of complex, real-world domains. In these environments, the ability to perceive and adapt to time stream, data stream, and model stream drifts is not merely a performance enhancement, but a prerequisite for sustained reliability. This section highlights several critical application domains where drift learning methodologies are actively bridging the gap between static models and lifelong self-learning (i.e., autonomous learning and decision-making without human intervention).

\subsection{Real Time Prediction}

Intelligent transportation systems, particularly train-carriage load prediction, represent a critical real-world application of drift learning in non-stationary environments. Accurately forecasting passenger distribution across train carriages is essential for managing platform crowding and optimizing boarding efficiency. Real-time carriage load data—often captured via on-board handlers such as weight sensors when doors close—forms a continuous data stream that is inherently subject to both rhythmic drift and arbitrary drift.

Furthermore, these systems face significant representational and structural heterogeneity because monitoring capabilities vary across the fleet, while newer models provide real-time load reporting, other train sets lack these sensors entirely. To bridge this gap and provide expected passenger loads across the entire network based on the time of day, researchers have leveraged multi-stream drift learning frameworks. For instance, Yu et al. \cite{yuRealTimeDecisionMaking2020} proposed a real-time decision-making framework utilizing multi-stream learning to capture evolving load dynamics. This was subsequently advanced into a multi-stream fuzzy learning system \cite{yuRealTimePredictionSystem2022}, which effectively models the structural correlations between different train lines. By continuously adapting to temporal variations and transferring predictive knowledge from sensor-rich to sensor-poor trains, these drift-aware autonomous learning systems ensure robust, network-wide passenger load forecasting.

\subsection{Recommender Systems}
Recommender systems are inherently susceptible to drift due to the continuous evolution of user preferences and item popularity within data streams. To maintain predictive accuracy in non-stationary environments, recent research has focused on enhancing the robustness and adaptability of recommendation models. For instance, the robustness of denoising recommendation frameworks has been rigorously evaluated to mitigate the impact of noisy user interactions \cite{zeng2025we}. To address the challenges of data sparsity and distributional shifts, cross-domain recommendation (CDR) strategies have been developed to facilitate knowledge transfer. In this context, the framework of AMT-CDR utilizes a deep adversarial multi-channel transfer network to align feature distributions across different domains \cite{lu2024amt}. Furthermore, a sharpness-aware optimization approach has been proposed for cross-domain recommendation to improve the generalization performance for cold-start users by seeking flatter minima in the loss landscape \cite{zeng2025sharpness}.

The integration of foundation models has further propelled the development of autonomous intelligence in recommendation. The system of RoSiLC-RS leverages LLM and step-back prompting to provide robust recommendations for similar legal cases, demonstrating superior reasoning capabilities in specialized domains \cite{zeng2025rosilc}. In the field of healthcare, personalized cancer risk prediction has been enhanced through LLM-driven recommender systems that incorporate genomics data to provide tailored medical insights \cite{lu2025genomics}. Additionally, novel learning paradigms such as partial-label learning have been introduced to handle the inherent ambiguity in user feedback, as exemplified by the framework of PARS \cite{ye2026pars}. These advancements reflect a strategic shift toward drift-aware and autonomous recommendation engines that can effectively navigate the complexities of dynamic data streams.

\subsection{Health and Pathology}

In the medical domain, diagnostic models frequently encounter severe representation and semantic drifts due to variations in patient demographics, hardware upgrades, and evolving clinical protocols. A prominent application is the analysis of gigapixel whole-slide images (WSIs) in digital pathology. When deploying VLMs or large convolutional networks across different hospitals, the visual characteristics of tissue slides can shift dramatically, a phenomenon that has been empirically shown to undermine the robustness of even state-of-the-art foundation models \cite{stacke2020measuring, thiringer2026scanner, chen2026tc}. Drift learning enables these systems to continuously align their feature representations against realistic domain shifts without requiring massive dataset retraining \cite{noori2026histopath, kumariattention}. Furthermore, autonomous adaptation frameworks allow these large models to explicitly detect and adapt to data-centric concept drift \cite{guan2026detecting}. By coupling these adaptive mechanisms with efficient structural strategies—such as token aggregation to accelerate computation over the massive spatial dimensions of WSIs—these systems can ensure that diagnostic accuracy remains robust even as the underlying medical imaging concepts evolve over time.

\subsection{Continuous Tuning of Multi-Modal Large Models}

As large multi-modal models (LMMs) are continuously updated to handle new tasks or domains, they inevitably experience model stream drift, particularly sequence and policy drifts. During non-stationary custom-tuning, a critical challenge is walking the tightrope between absorbing new, task-specific knowledge and retaining foundational reasoning capabilities \cite{yang2025walking}. Drift learning applications in this space focus on autonomously disentangling beneficial drifts—which improve alignment with the new target distribution—from detrimental drifts that cause catastrophic forgetting or semantic collapse \cite{yang2025learning,yang2025resilient}. This ensures that LMMs can safely evolve and specialize in decentralized or federated settings without degrading their core competencies.

\subsection{Autonomous Vehicles and Robotics}

Autonomous navigation presents one of the most demanding applications for drift learning, as vehicles operate in highly volatile and open-world environments. Perception and planning modules face continuous arbitrary drifts due to sudden weather changes, varying lighting conditions, and unpredictable pedestrian behaviors, which frequently degrade the performance of stationary models \cite{sun2022shift, li2024domain}. To counteract this, reinforcement learning agents driving these vehicles must employ policy constraints and off-policy rectification to manage policy drift. Recent constrained adversarial reinforcement learning frameworks have demonstrated that bounding policy updates ensures the vehicle's decision-making remains safe even when the target policy deviates from the behavior policy under environmental perturbations \cite{xu2026vgas}. Furthermore, by integrating proactive drift anticipation and temporal consistency mechanisms, autonomous ground robots can seamlessly transition between domains while maintaining operational stability and safety \cite{gu2025climb}.

\section{Conclusion and Future Directions}

This survey has presented a comprehensive framework for Drift Learning, redefining it as a foundational paradigm for achieving self-learning (i.e., autonomous learning and decision-making without human intervention) in open and non-stationary environments.
Moving beyond the traditional view of concept drift as a mere nuisance to be detected and corrected, we have systematized the field through three interconnected dimensions: time stream drift, data stream drift, and model stream drift.
This taxonomy provides a holistic lens to understand how learning systems perceive, interpret, and evolve amidst change.

First, by analyzing time stream Drift, we distinguished between Arbitrary Drift, which demands rapid reactive detection, and Rhythmic Drift, which offers opportunities for proactive forecasting and anticipation.
This distinction highlights a critical shift from passive error-correction to temporal structural modeling, enabling systems to exploit inherent regularities in dynamic streams.
Second, our exploration of data stream drift decoupled the evolution of feature distributions (Representation Drift) from the shifts in underlying meaning (Semantic Drift).
We underscored that robust adaptation requires not only aligning statistical discrepancies but also preserving semantic consistency through invariant learning and generative reconstruction.
Third, the formulation of model stream Drift addressed the internal dynamics of the learning agent itself.
Whether managing the stability-plasticity dilemma in sequential updates (Sequence Drift), reconciling divergent parameters in decentralized networks (Heterogeneity Drift), or stabilizing feedback loops in decision-making (Policy Drift), the core challenge lies in maintaining a dynamic equilibrium during continuous evolution.

In conclusion, drift learning represents the bridge between static model optimization and lifelong autonomous evolution.
Future research must move towards integrated architectures that can simultaneously disentangle complex drift sources, reason about their causality, and adapt in a self-sustaining manner.
By endowing machines with the ability to learn through change rather than merely despite it, we pave the way for truly resilient and adaptive artificial intelligence.

\section*{Acknowledgment}
The work was supported by the Australian Research Council (ARC) under Laureate project FL190100149.

\bibliographystyle{IEEEtran}
\bibliography{new}

@article{yang2026towards,
  title={Towards Robust Endogenous Reasoning: Unifying Drift Adaptation in Non-Stationary Tuning},
  author={Yang, Xiaoyu and Yu, En and Duan, Wei and Lu, Jie},
  journal={arXiv preprint arXiv:2604.15705},
  year={2026}
}

@inproceedings{yangAdaptingMultimodalLarge2024b,
  author = {Yang, Xiaoyu and Lu, Jie and Yu, En},
  booktitle={The Thirteenth International Conference on Learning Representations},
  editor = {Y. Yue and A. Garg and N. Peng and F. Sha and R. Yu},
  pages = {90869--90891},
  title = {Adapting Multi-modal Large Language Model to Concept Drift From Pre-training Onwards},
  url = {https://proceedings.iclr.cc/paper_files/paper/2025/file/e25d87b8a42ee3f0d5b3ef741ca13031-Paper-Conference.pdf},
  volume = {2025},
  year = {2025}
}

@inproceedings{yang2025walking,
  title={Walking the Tightrope: Autonomous Disentangling Beneficial and Detrimental Drifts in Non-Stationary Custom-Tuning},
  author={Yang, Xiaoyu and Lu, Jie and Yu, En},
  booktitle={The Thirty-ninth Annual Conference on Neural Information Processing Systems},
  year = {2025},
  url={https://openreview.net/forum?id=1BAiQmAFsx}  
}

@article{yangTdistributedSphericalFeature2023,
  title = {T-Distributed {{Spherical Feature Representation}} for {{Imbalanced Classification}}},
  author = {Yang, Xiaoyu and Chen, Yufei and Yue, Xiaodong and Xu, Shaoxun and Ma, Chao},
  year = {2023},
  month = jun,
  journal = {Proceedings of the AAAI Conference on Artificial Intelligence},
  volume = {37},
  number = {9},
  pages = {10825--10833},
  issn = {2374-3468},
  doi = {10.1609/aaai.v37i9.26284},
  copyright = {Copyright (c) 2023 Association for the Advancement of Artificial Intelligence},
  langid = {english}
}

@InProceedings{yangone,
  title = {One Leaf Reveals the Season: Occlusion-Based Contrastive Learning with Semantic-Aware Views for Efficient Visual Representation},
  author = {Yang, Xiaoyu and Xu, Lijian and Li, Hongsheng and Zhang, Shaoting},
  booktitle = {Proceedings of the 42nd International Conference on Machine Learning},
  pages = {71425--71440},
  year = {2025},
  editor = {Singh, Aarti and Fazel, Maryam and Hsu, Daniel and Lacoste-Julien, Simon and Berkenkamp, Felix and Maharaj, Tegan and Wagstaff, Kiri and Zhu, Jerry},
  volume = {267},
  series = {Proceedings of Machine Learning Research},
  month = {13--19 Jul},
  publisher = {PMLR},
  url = {https://proceedings.mlr.press/v267/yang25ao.html}
}

@article{yang2024segmentation,
  author={Yang, Xiaoyu and Xu, Lijian and Yu, Simon and Xia, Qing and Li, Hongsheng and Zhang, Shaoting},
  journal={IEEE Transactions on Medical Imaging}, 
  title={Segmentation and Vascular Vectorization for Coronary Artery by Geometry-Based Cascaded Neural Network}, 
  year={2025},
  volume={44},
  number={1},
  pages={259-269},
  doi={10.1109/TMI.2024.3435714}
}

@article{yang2022local,
  title={Local linear embedding based interpolation neural network in pancreatic tumor segmentation},
  author={Yang, Xiaoyu and Chen, Yufei and Yue, Xiaodong and Ma, Chao and Yang, Panpan},
  journal={Applied Intelligence},
  volume={52},
  number={8},
  pages={8746--8756},
  year={2022},
  publisher={Springer}
}

@article{young2026xrayclaw,
  title={XrayClaw: Cooperative-Competitive Multi-Agent Alignment for Trustworthy Chest X-ray Diagnosis},
  author={Young, Shawn and Xu, Lijian},
  journal={arXiv preprint arXiv:2604.02695},
  year={2026}
}

@article{yang2025learning,
  title={Learning from All: Concept Alignment for Autonomous Distillation from Multiple Drifting MLLMs},
  author={Yang, Xiaoyu and Lu, Jie and Yu, En},
  journal={arXiv preprint arXiv:2510.04142},
  year={2025}
}

@article{yang2025resilient,
  title={Resilient Contrastive Pre-training under Non-Stationary Drift},
  author={Yang, Xiaoyu and Lu, Jie and Yu, En and Duan, Wei},
  journal={arXiv preprint arXiv:2502.07620},
  year={2025}
}

@article{yang2026scalar,
  title={SCALAR: Spatial-Concept Alignment for Robust Vision in Harsh Open World},
  author={Yang, Xiaoyu and Xu, Lijian and Zeng, Xingyu and Wang, Xiaosong and Li, Hongsheng and Zhang, Shaoting},
  journal={Pattern Recognition},
  pages={113203},
  year={2026},
  publisher={Elsevier}
}

@article{chen2026tc,
  title={TC-SSA: Token Compression via Semantic Slot Aggregation for Gigapixel Pathology Reasoning},
  author={Chen, Zhuo and Young, Shawn and Xu, Lijian},
  journal={arXiv preprint arXiv:2603.01143},
  year={2026}
}

@article{young2025fewer,
  title={Fewer Tokens, Greater Scaling: Self-Adaptive Visual Bases for Efficient and Expansive Representation Learning},
  author={Young, Shawn and Zeng, Xingyu and Xu, Lijian},
  journal={arXiv preprint arXiv:2511.19515},
  year={2025}
}

@inproceedings{yu2025learning,
  title={Learning Robust Spectral Dynamics for Temporal Domain Generalization},
  author={Yu, En and Lu, Jie and Yang, Xiaoyu and Zhang, Guangquan and Fang, Zhen},
  booktitle={The Thirty-ninth Annual Conference on Neural Information Processing Systems},
  year={2025},
}

@inproceedings{yu2025drift,
  title={Drift-aware collaborative assistance mixture of experts for heterogeneous multistream learning},
  author={Yu, En and Lu, Jie and Wang, Kun and Yang, Xiaoyu and Zhang, Guangquan},
  booktitle={Proceedings of the AAAI Conference on Artificial Intelligence},
  volume={40},
  number={19},
  pages={16199--16207},
  year={2026}
}

@article{yuDetectingGroupConcept2023c,
  title = {Detecting Group Concept Drift from Multiple Data Streams},
  author = {Yu, Hang and Liu, Weixu and Lu, Jie and Wen, Yimin and Luo, Xiangfeng and Zhang, Guangquan},
  year = 2023,
  month = feb,
  journal = {Pattern Recognition},
  volume = {134},
  pages = {109113},
  issn = {0031-3203},
  doi = {10.1016/j.patcog.2022.109113},
  urldate = {2025-10-25}
}

@article{yu2026autonomous,
  author={Yu, En and Lu, Jie and Yang, Xiaoyu and Zhang, Guangquan},
  journal={IEEE Transactions on Fuzzy Systems}, 
  title={Autonomous Online Multistream Generalization via Fuzzy Joint Discriminant Analysis}, 
  year={2026},
  volume={34},
  number={4},
  pages={1256-1268},
  doi={10.1109/TFUZZ.2026.3656402}}

@inproceedings{kalbPrinciplesForgettingDomainIncremental2023,
  title = {Principles of {{Forgetting}} in {{Domain-Incremental Semantic Segmentation}} in {{Adverse Weather Conditions}}},
  booktitle = {Proceedings of the {{IEEE}}/{{CVF Conference}} on {{Computer Vision}} and {{Pattern Recognition}}},
  author = {Kalb, Tobias and Beyerer, J{\"u}rgen},
  year = {2023},
  pages = {19508--19518},
  urldate = {2025-10-09},
  langid = {english}
}

@article{chenAlgorithmicFairnessArtificial2023,
  title = {Algorithmic Fairness in Artificial Intelligence for Medicine and Healthcare},
  author = {Chen, Richard J. and Wang, Judy J. and Williamson, Drew F. K. and Chen, Tiffany Y. and Lipkova, Jana and Lu, Ming Y. and Sahai, Sharifa and Mahmood, Faisal},
  year = {2023},
  month = jun,
  journal = {Nature Biomedical Engineering},
  volume = {7},
  number = {6},
  pages = {719--742},
  publisher = {Nature Publishing Group},
  issn = {2157-846X},
  doi = {10.1038/s41551-023-01056-8},
  urldate = {2025-10-09},
  copyright = {2023 Springer Nature Limited},
  langid = {english}
}

@inproceedings{baena2006early,
  title={Early drift detection method},
  author={Baena-Garc{\i}a, Manuel and del Campo-{\'A}vila, Jos{\'e} and Fidalgo, Ra{\'u}l and Bifet, Albert and Gavalda, R and Morales-Bueno, R},
  booktitle={Fourth international workshop on knowledge discovery from data streams},
  volume={6},
  pages={77--86},
  year={2006}
}

@inproceedings{du2019multi,
  title={Multi-source transfer learning for non-stationary environments},
  author={Du, Honghui and Minku, Leandro L and Zhou, Huiyu},
  booktitle={2019 International Joint Conference on Neural Networks (IJCNN)},
  pages={1--8},
  year={2019},
  organization={IEEE}
}

@article{yu2024fuzzy,
  author={Yu, En and Lu, Jie and Zhang, Guangquan},
  journal={IEEE Transactions on Fuzzy Systems}, 
  title={Fuzzy Shared Representation Learning for Multistream Classification}, 
  year={2024},
  volume={32},
  number={10},
  pages={5625-5637}}

@article{renchunzi2022automatic,
  title={Automatic online multi-source domain adaptation},
  author={Renchunzi, Xie and Pratama, Mahardhika},
  journal={Information Sciences},
  volume={582},
  pages={480--494},
  year={2022},
  publisher={Elsevier}
}

@article{jiao2022reduced,
  title={Reduced-space Multistream Classification based on Multi-objective Evolutionary Optimization},
  author={Jiao, Botao and Guo, Yinan and Yang, Shengxiang and Pu, Jiayang and Gong, Dunwei},
  journal={IEEE Transactions on Evolutionary Computation},
  year={2022},
  publisher={IEEE}
}

@article{yu2022meta,
  title={Meta-ADD: A meta-learning based pre-trained model for concept drift active detection},
  author={Yu, Hang and Zhang, Qingyong and Liu, Tianyu and Lu, Jie and Wen, Yimin and Zhang, Guangquan},
  journal={Information Sciences},
  volume={608},
  pages={996--1009},
  year={2022},
  publisher={Elsevier}
}

@article{yu2022learn,
  title={Learn-to-adapt: Concept drift adaptation for hybrid multiple streams},
  author={Yu, En and Song, Yiliao and Zhang, Guangquan and Lu, Jie},
  journal={Neurocomputing},
  volume={496},
  pages={121--130},
  year={2022},
  publisher={Elsevier}
}

@inproceedings{haque2017fusion,
  title={Fusion: An online method for multistream classification},
  author={Haque, Ahsanul and Wang, Zhuoyi and Chandra, Swarup and Dong, Bo and Khan, Latifur and Hamlen, Kevin W},
  booktitle={Proceedings of the 2017 ACM on Conference on Information and Knowledge Management},
  pages={919--928},
  year={2017}
}

@inproceedings{pratama2019atl,
  title={ATL: Autonomous knowledge transfer from many streaming processes},
  author={Pratama, Mahardhika and de Carvalho, Marcus and Xie, Renchunzi and Lughofer, Edwin and Lu, Jie},
  booktitle={Proceedings of the 28th ACM International Conference on Information and Knowledge Management},
  pages={269--278},
  year={2019}
}

@article{xu2025coral,
  title={Coral: Concept drift representation learning for co-evolving time-series},
  author={Xu, Kunpeng and Chen, Lifei and Wang, Shengrui},
  journal={arXiv preprint arXiv:2501.01480},
  year={2025}
}

@article{wen2023onenet,
  title={Onenet: Enhancing time series forecasting models under concept drift by online ensembling},
  author={Wen, Qingsong and Chen, Weiqi and Sun, Liang and Zhang, Zhang and Wang, Liang and Jin, Rong and Tan, Tieniu and others},
  journal={Advances in Neural Information Processing Systems},
  volume={36},
  pages={69949--69980},
  year={2023}
}

@inproceedings{wang2024adaptive,
  title={An Adaptive Stacking Method for Multiple Data Streams Learning under Concept Drift},
  author={Wang, Kun and Lu, Jie and Liu, Anjin and Zhang, Guangquan},
  booktitle={The 19th ISKE Conference on Intelligence Systems and Knowledge Engineering (FLINS-ISKE 2024)},
  pages={267--274},
  year={2024},
  organization={World Scientific}
}

@article{xu2025drift2matrix,
  title = {CORAL: Concept Drift Representation Learning for Co-evolving Time-series},
  author = {Xu, Kunpeng and Chen, Lifei and Wang, Shengrui},
  journal = {arXiv preprint arXiv:2501.01480},
  year = {2025},
}

@inproceedings{wang2020continuously,
  title={Continuously indexed domain adaptation},
  author={Wang, Hao and He, Hao and Katabi, Dina},
  booktitle={Proceedings of the 37th International Conference on Machine Learning},
  pages={9898--9907},
  year={2020}
}

@inproceedings{chang2023coda,
  title={Coda: Temporal domain generalization via concept drift simulator},
  author={Chang, Chia-Yuan and Chuang, Yu-Neng and Jiang, Zhimeng and Lai, Kwei-Herng and Jiang, Anxiao and Zou, Na},
  booktitle={Proceedings of the 31st ACM SIGKDD Conference on Knowledge Discovery and Data Mining V. 2},
  pages={131--142},
  year={2025}
}

@article{cai2024continuous,
  title={Continuous temporal domain generalization},
  author={Cai, Zekun and Bai, Guangji and Jiang, Renhe and Song, Xuan and Zhao, Liang},
  journal={Advances in Neural Information Processing Systems},
  volume={37},
  pages={127987--128014},
  year={2024}
}

@inproceedings{li2022ddg,
  title={Ddg-da: Data distribution generation for predictable concept drift adaptation},
  author={Li, Wendi and Yang, Xiao and Liu, Weiqing and Xia, Yingce and Bian, Jiang},
  booktitle={Proceedings of the AAAI Conference on Artificial Intelligence},
  volume={36},
  number={4},
  pages={4092--4100},
  year={2022}
}

@article{abdullahi2025systematic,
  title={A Systematic Literature Review of Concept Drift Mitigation in Time-Series Applications},
  author={Abdullahi, Mujaheed and Alhussian, Hitham and Aziz, Norshakirah and Abdulkadir, Said Jadid and Baashar, Yahia and Alashhab, Abdussalam Ahmed and Afrin, Afroza},
  journal={IEEE Access},
  year={2025},
  publisher={IEEE}
}

@inproceedings{zhao2025proactive,
  title={Proactive model adaptation against concept drift for online time series forecasting},
  author={Zhao, Lifan and Shen, Yanyan},
  booktitle={Proceedings of the 31st ACM SIGKDD Conference on Knowledge Discovery and Data Mining V. 1},
  pages={2020--2031},
  year={2025}
}

@inproceedings{xu2024kan4drift,
  title={Kan4drift: Are kan effective for identifying and tracking concept drift in time series?},
  author={Xu, Kunpeng and Chen, Lifei and Wang, Shengrui},
  booktitle={NeurIPS Workshop on Time Series in the Age of Large Models},
  year={2024}
}

@inproceedings{zhou2022fedformer,
  title={Fedformer: Frequency enhanced decomposed transformer for long-term series forecasting},
  author={Zhou, Tian and Ma, Ziqing and Wen, Qingsong and Wang, Xue and Sun, Liang and Jin, Rong},
  booktitle={International conference on machine learning},
  pages={27268--27286},
  year={2022},
  organization={PMLR}
}

@inproceedings{
ye2024frequency,
title={Frequency Adaptive Normalization For Non-stationary Time Series Forecasting},
author={Weiwei Ye and Songgaojun Deng and Qiaosha Zou and Ning Gui},
booktitle={The Thirty-eighth Annual Conference on Neural Information Processing Systems},
year={2024},
url={https://openreview.net/forum?id=T0axIflVDD}
}

@inproceedings{li2023atfnet,
  title={Fedformer: Frequency enhanced decomposed transformer for long-term series forecasting},
  author={Zhou, Tian and Ma, Ziqing and Wen, Qingsong and Wang, Xue and Sun, Liang and Jin, Rong},
  booktitle={International conference on machine learning},
  pages={27268--27286},
  year={2022},
  organization={PMLR}
}

@article{liu2023koopa,
  title={Koopa: Learning non-stationary time series dynamics with koopman predictors},
  author={Liu, Yong and Li, Chenyu and Wang, Jianmin and Long, Mingsheng},
  journal={Advances in neural information processing systems},
  volume={36},
  pages={12271--12290},
  year={2023}
}

@inproceedings{
liu2023adaptive,
title={Adaptive Normalization for Non-stationary Time Series Forecasting: A Temporal Slice Perspective},
author={Zhiding Liu and Mingyue Cheng and Zhi Li and Zhenya Huang and Qi Liu and Yanhu Xie and Enhong Chen},
booktitle={Thirty-seventh Conference on Neural Information Processing Systems},
year={2024},
url={https://openreview.net/forum?id=5BqDSw8r5j}
}

@inproceedings{you2021learning,
  title={Learning to learn the future: Modeling concept drifts in time series prediction},
  author={You, Xiaoyu and Zhang, Mi and Ding, Daizong and Feng, Fuli and Huang, Yuanmin},
  booktitle={Proceedings of the 30th ACM International Conference on Information \& Knowledge Management},
  pages={2434--2443},
  year={2021}
}

@article{zhang2025learning,
  title={Learning Unbiased Cluster Descriptors for Interpretable Imbalanced Concept Drift Detection},
  author={Zhang, Yiqun and Huang, Zhanpei and Zhao, Mingjie and Zhang, Chuyao and Lu, Yang and Ji, Yuzhu and Gu, Fangqing and Zeng, An},
  journal={IEEE Transactions on Emerging Topics in Computational Intelligence},
  year={2025},
  publisher={IEEE}
}

@article{greco2025unsupervised,
  title={Unsupervised concept drift detection from deep learning representations in real-time},
  author={Greco, Salvatore and Vacchetti, Bartolomeo and Apiletti, Daniele and Cerquitelli, Tania},
  journal={IEEE Transactions on Knowledge and Data Engineering},
  year={2025},
  publisher={IEEE}
}

@article{hou2025osasformer,
  title={OSASformer: A transformer-based model for OSAS screening via multi-source representation fusion},
  author={Hou, Yuanyuan and Wang, Bin and Zhang, Chengxi and Wang, Qiang and Li, Jiang and Meng, Pingping and Zhang, Yongxiang and Han, Chao and Hong, Feng and Zhang, Tong},
  journal={Knowledge-Based Systems},
  volume={316},
  pages={113365},
  year={2025},
  publisher={Elsevier}
}

@article{xie2024evolving,
  title={Evolving standardization for continual domain generalization over temporal drift},
  author={Xie, Mixue and Li, Shuang and Yuan, Longhui and Liu, Chi and Dai, Zehui},
  journal={Advances in Neural Information Processing Systems},
  volume={36},
  year={2024}
}

@article{qin2023evolving,
  title={Evolving domain generalization via latent structure-aware sequential autoencoder},
  author={Qin, Tiexin and Wang, Shiqi and Li, Haoliang},
  journal={IEEE Transactions on Pattern Analysis and Machine Intelligence},
  volume={45},
  number={12},
  pages={14514--14527},
  year={2023},
  publisher={IEEE}
}

@article{he2025learning,
  title={Learning Time-Aware Causal Representation for Model Generalization in Evolving Domains},
  author={He, Zhuo and Li, Shuang and Song, Wenze and Yuan, Longhui and Liang, Jian and Li, Han and Gai, Kun},
  journal={arXiv preprint arXiv:2506.17718},
  year={2025}
}

@inproceedings{lu2025early,
  title={Early concept drift detection via prediction uncertainty},
  author={Lu, Pengqian and Lu, Jie and Liu, Anjin and Zhang, Guangquan},
  booktitle={Proceedings of the AAAI Conference on Artificial Intelligence},
  volume={39},
  number={18},
  pages={19124--19132},
  year={2025}
}

@article{lu2025autonomous,
  title={Autonomous Concept Drift Threshold Determination},
  author={Lu, Pengqian and Lu, Jie and Liu, Anjin and Yu, En and Zhang, Guangquan},
  journal={arXiv preprint arXiv:2511.09953},
  year={2025}
}

@article{nasery2021training,
  title={Training for the future: A simple gradient interpolation loss to generalize along time},
  author={Nasery, Anshul and Thakur, Soumyadeep and Piratla, Vihari and De, Abir and Sarawagi, Sunita},
  journal={Advances in Neural Information Processing Systems},
  volume={34},
  pages={19198--19209},
  year={2021}
}

@inproceedings{hu2020domain,
  title={Domain generalization via multidomain discriminant analysis},
  author={Hu, Shoubo and Zhang, Kun and Chen, Zhitang and Chan, Laiwan},
  booktitle={Uncertainty in artificial intelligence},
  pages={292--302},
  year={2020},
  organization={PMLR}
}

@article{lu2018learning,
  title={Learning under concept drift: A review},
  author={Lu, Jie and Liu, Anjin and Dong, Fan and Gu, Feng and Gama, Joao and Zhang, Guangquan},
  journal={IEEE Transactions on Knowledge and Data Engineering},
  volume={31},
  number={12},
  pages={2346--2363},
  year={2018},
  publisher={IEEE}
}

@inproceedings{gama2004learning,
  title={Learning with drift detection},
  author={Gama, Joao and Medas, Pedro and Castillo, Gladys and Rodrigues, Pedro},
  booktitle={Brazilian Symposium on Artificial Intelligence},
  pages={286--295},
  year={2004},
  organization={Springer}
}

@article{frias2014online,
  title={Online and non-parametric drift detection methods based on Hoeffding’s bounds},
  author={Frias-Blanco, Isvani and del Campo-{\'A}vila, Jos{\'e} and Ramos-Jimenez, Gonzalo and Morales-Bueno, Rafael and Ortiz-Diaz, Agustin and Caballero-Mota, Yail{\'e}},
  journal={IEEE Transactions on Knowledge and Data Engineering},
  volume={27},
  number={3},
  pages={810--823},
  year={2014},
  publisher={IEEE}
}

@inproceedings{bifet2007learning,
  title={Learning from time-changing data with adaptive windowing},
  author={Bifet, Albert and Gavalda, Ricard},
  booktitle={Proceedings of the 2007 SIAM International Conference on Data Mining},
  pages={443--448},
  year={2007},
  organization={SIAM}
}

@article{alippi2008just,
  title={Just-in-time adaptive classifiers—Part I: Detecting nonstationary changes},
  author={Alippi, Cesare and Roveri, Manuel},
  journal={IEEE Transactions on Neural Networks},
  volume={19},
  number={7},
  pages={1145--1153},
  year={2008},
  publisher={IEEE}
}

@inproceedings{yu2017concept,
  title={Concept drift detection with hierarchical hypothesis testing},
  author={Yu, Shujian and Abraham, Zubin},
  booktitle={Proceedings of the 2017 SIAM International Conference on Data Mining},
  pages={768--776},
  year={2017},
  organization={SIAM}
}

@inproceedings{bach2008paired,
  title={Paired learners for concept drift},
  author={Bach, Stephen H and Maloof, Marcus A},
  booktitle={2008 Eighth IEEE International Conference on Data Mining},
  pages={23--32},
  year={2008},
  organization={IEEE}
}

@article{liu2016fp,
  title={FP-ELM: An online sequential learning algorithm for dealing with concept drift},
  author={Liu, Dong and Wu, YouXi and Jiang, He},
  journal={Neurocomputing},
  volume={207},
  pages={322--334},
  year={2016},
  publisher={Elsevier}
}

@inproceedings{chenDelvingTrajectoryLongtail2024,
  title = {Delving into the {{Trajectory Long-tail Distribution}} for {{Muti-object Tracking}}},
  booktitle = {Proceedings of the {{IEEE}}/{{CVF Conference}} on {{Computer Vision}} and {{Pattern Recognition}}},
  author = {Chen, Sijia and Yu, En and Li, Jinyang and Tao, Wenbing},
  year = {2024},
  pages = {19341--19351},
  langid = {english}
}

@inproceedings{yu2024online,
  title={Online boosting adaptive learning under concept drift for multistream classification},
  author={Yu, En and Lu, Jie and Zhang, Bin and Zhang, Guangquan},
  booktitle={Proceedings of the AAAI Conference on Artificial Intelligence},
  volume={38},
  number={15},
  pages={16522--16530},
  year={2024}
}

@article{wang2025Adaptive,
  author={Wang, Kun and Lu, Jie and Liu, Anjin},
  journal={IEEE Transactions on Knowledge and Data Engineering}, 
  title={Adaptive Information Fusion-Based Concept Drift Learning for Evolving Multiple Data Streams}, 
  year={2025},
  volume={37},
  number={12},
  pages={6863-6876}}

@article{zhou2025continuous,
  title={Continuous Graph Learning-Based Self-Adaptation for Multi-Stream Concept Drift},
  author={Zhou, Ming and Lu, Jie},
  journal={IEEE Transactions on Cybernetics},
  year={2025},
  publisher={IEEE}
}

@article{zhou2023multi,
  title={Multi-stream concept drift self-adaptation using graph neural network},
  author={Zhou, Ming and Lu, Jie and Song, Yiliao and Zhang, Guangquan},
  journal={IEEE Transactions on Knowledge and Data Engineering},
  volume={35},
  number={12},
  pages={12828--12841},
  year={2023},
  publisher={IEEE}
}

@article{zhang2025tracking,
  title={Tracking Correlations Between Multiple Data Streams Through Evolutionary Regressor Chains},
  author={Zhang, Bin and Lu, Jie and Liu, Anjin and Yao, Xin and Zhang, Guangquan},
  journal={IEEE Transactions on Cybernetics},
  year={2025},
  publisher={IEEE}
}

@article{zhang2025multistream,
  title={A Multistream Concept Drift Handling Framework via Data Sharing},
  author={Zhang, Bin and Lu, Jie and Song, Yiliao and Zhang, Guangquan},
  journal={IEEE Transactions on Cybernetics},
  year={2025},
  publisher={IEEE}
}

@inproceedings{hoLongTailedAnomalyDetection2024,
  title = {Long-{{Tailed Anomaly Detection}} with {{Learnable Class Names}}},
  booktitle = {Proceedings of the {{IEEE}}/{{CVF Conference}} on {{Computer Vision}} and {{Pattern Recognition}}},
  author = {Ho, Chih-Hui and Peng, Kuan-Chuan and Vasconcelos, Nuno},
  year = {2024},
  pages = {12435--12446},
  langid = {english}
}

@inproceedings{parasharNeglectedTailsVisionLanguage2024,
  title = {The {{Neglected Tails}} in {{Vision-Language Models}}},
  booktitle = {Proceedings of the {{IEEE}}/{{CVF Conference}} on {{Computer Vision}} and {{Pattern Recognition}}},
  author = {Parashar, Shubham and Lin, Zhiqiu and Liu, Tian and Dong, Xiangjue and Li, Yanan and Ramanan, Deva and Caverlee, James and Kong, Shu},
  year = {2024},
  pages = {12988--12997},
  langid = {english}
}

@article{gomes2017adaptive,
  title={Adaptive random forests for evolving data stream classification},
  author={Gomes, Heitor M and Bifet, Albert and Read, Jesse and Barddal, Jean Paul and Enembreck, Fabr{\'\i}cio and Pfharinger, Bernhard and Holmes, Geoff and Abdessalem, Talel},
  journal={Machine Learning},
  volume={106},
  number={9},
  pages={1469--1495},
  year={2017},
  publisher={Springer}
}

@article{kolter2007dynamic,
  title={Dynamic weighted majority: An ensemble method for drifting concepts},
  author={Kolter, J Zico and Maloof, Marcus A},
  journal={The Journal of Machine Learning Research},
  volume={8},
  pages={2755--2790},
  year={2007},
  publisher={JMLR.org}
}

@article{xu2017dynamic,
  title={Dynamic extreme learning machine for data stream classification},
  author={Xu, Shuliang and Wang, Junhong},
  journal={Neurocomputing},
  volume={238},
  pages={433--449},
  year={2017},
  publisher={Elsevier}
}

@inproceedings{domingos2000mining,
  title={Mining high-speed data streams},
  author={Domingos, Pedro and Hulten, Geoff},
  booktitle={Proceedings of the sixth ACM SIGKDD International Conference on Knowledge Discovery and Data Mining},
  pages={71--80},
  year={2000}
}

@inproceedings{rangwaniDeiTLTDistillationStrikes2024,
  title = {{{DeiT-LT}}: {{Distillation Strikes Back}} for {{Vision Transformer Training}} on {{Long-Tailed Datasets}}},
  shorttitle = {{{DeiT-LT}}},
  booktitle = {Proceedings of the {{IEEE}}/{{CVF Conference}} on {{Computer Vision}} and {{Pattern Recognition}}},
  author = {Rangwani, Harsh and Mondal, Pradipto and Mishra, Mayank and Asokan, Ashish Ramayee and Babu, R. Venkatesh},
  year = {2024},
  pages = {23396--23406},
  langid = {english}
}

@inproceedings{wangLongTailClassIncremental2024,
  title = {Long-{{Tail Class Incremental Learning}} via {{Independent Sub-prototype Construction}}},
  booktitle = {Proceedings of the {{IEEE}}/{{CVF Conference}} on {{Computer Vision}} and {{Pattern Recognition}}},
  author = {Wang, Xi and Yang, Xu and Yin, Jie and Wei, Kun and Deng, Cheng},
  year = {2024},
  pages = {28598--28607},
  langid = {english}
}

@inproceedings{zhengBEMBalancedEntropybased2024,
  title = {{{BEM}}: {{Balanced}} and {{Entropy-based Mix}} for {{Long-Tailed Semi-Supervised Learning}}},
  shorttitle = {{{BEM}}},
  booktitle = {Proceedings of the {{IEEE}}/{{CVF Conference}} on {{Computer Vision}} and {{Pattern Recognition}}},
  author = {Zheng, Hongwei and Zhou, Linyuan and Li, Han and Su, Jinming and Wei, Xiaoming and Xu, Xiaoming},
  year = {2024},
  pages = {22893--22903},
  langid = {english}
}

@inproceedings{jungTailedCoreFewShotSampling2025,
  title = {{{TailedCore}}: {{Few-Shot Sampling}} for {{Unsupervised Long-Tail Noisy Anomaly Detection}}},
  shorttitle = {{{TailedCore}}},
  booktitle = {Proceedings of the {{Computer Vision}} and {{Pattern Recognition Conference}}},
  author = {Jung, Yoon Gyo and Park, Jaewoo and Yoon, Jaeho and Peng, Kuan-Chuan and Kim, Wonchul and Teoh, Andrew Beng Jin and Camps, Octavia},
  year = {2025},
  pages = {25539--25548},
  langid = {english}
}

@inproceedings{sidhuSearchDetectTrainingFree2025,
  title = {Search and {{Detect}}: {{Training-Free Long Tail Object Detection}} via {{Web-Image Retrieval}}},
  shorttitle = {Search and {{Detect}}},
  booktitle = {Proceedings of the {{Computer Vision}} and {{Pattern Recognition Conference}}},
  author = {Sidhu, Mankeerat and Chopra, Hetarth and Blume, Ansel and Kim, Jeonghwan and Reddy, Revanth Gangi and Ji, Heng},
  year = {2025},
  pages = {15129--15138},
  langid = {english}
}

@article{duProbabilisticContrastiveLearning2024,
  title = {Probabilistic {{Contrastive Learning}} for {{Long-Tailed Visual Recognition}}},
  author = {Du, Chaoqun and Wang, Yulin and Song, Shiji and Huang, Gao},
  year = {2024},
  month = sep,
  journal = {IEEE Transactions on Pattern Analysis and Machine Intelligence},
  volume = {46},
  number = {9},
  pages = {5890--5904},
  issn = {1939-3539},
  doi = {10.1109/TPAMI.2024.3369102},
  langid = {american}
}

@inproceedings{sunRethinkingClassifierReTraining2024,
  title = {Rethinking {{Classifier Re-Training}} in {{Long-Tailed Recognition}}: {{Label Over-Smooth Can Balance}}},
  shorttitle = {Rethinking {{Classifier Re-Training}} in {{Long-Tailed Recognition}}},
  booktitle = {The {{Thirteenth International Conference}} on {{Learning Representations}}},
  author = {Sun, Siyu and Lu, Han and Li, Jiangtong and Xie, Yichen and Li, Tianjiao and Yang, Xiaokang and Zhang, Liqing and Yan, Junchi},
  year = {2024},
  langid = {english}
}

@inproceedings{wangKillTwoBirds2023,
  title = {Kill {{Two Birds}} with {{One Stone}}: {{Rethinking Data Augmentation}} for {{Deep Long-tailed Learning}}},
  shorttitle = {Kill {{Two Birds}} with {{One Stone}}},
  booktitle = {The {{Twelfth International Conference}} on {{Learning Representations}}},
  author = {Wang, Binwu and Wang, Pengkun and Xu, Wei and Wang, Xu and Zhang, Yudong and Wang, Kun and Wang, Yang},
  year = {2023},
  langid = {english}
}

@article{zhouContinuousContrastiveLearning2024,
  title = {Continuous {{Contrastive Learning}} for {{Long-Tailed Semi-Supervised Recognition}}},
  author = {Zhou, Zi-Hao and Fang, Siyuan and Zhou, Zi-Jing and Wei, Tong and Wan, Yuanyu and Zhang, Min-Ling},
  year = {2024},
  month = dec,
  journal = {Advances in Neural Information Processing Systems},
  volume = {37},
  pages = {51411--51435},
  langid = {english}
}

@article{dengEIFAKDExplicitImplicit2026,
  title = {{{EIFA-KD}}: {{Explicit}} and Implicit Feature Augmentation with Knowledge Distillation for Long-Tailed Visual Data Classification},
  shorttitle = {{{EIFA-KD}}},
  author = {Deng, Xiyan and Wang, Xiaoli and Sun, Yifan and Zhao, Xusheng and Tian, Siju and Li, Minqi and Wang, Yuping},
  year = {2026},
  month = mar,
  journal = {Pattern Recognition},
  volume = {171},
  pages = {112129},
  issn = {0031-3203},
  doi = {10.1016/j.patcog.2025.112129}
}

@article{guoPrototypeAlignmentDedicated2025,
  title = {Prototype {{Alignment With Dedicated Experts}} for {{Test-Agnostic Long-Tailed Recognition}}},
  author = {Guo, Chen and Chen, Weiling and Huang, Aiping and Zhao, Tiesong},
  year = {2025},
  journal = {IEEE Transactions on Multimedia},
  volume = {27},
  pages = {455--465},
  issn = {1941-0077},
  doi = {10.1109/TMM.2024.3521665}
}

@article{liSynthesizingMinoritySamples2025,
  title = {Synthesizing {{Minority Samples}} for {{Long-tailed Classification}} via {{Distribution Matching}}},
  author = {Li, Zhuo and Zhao, He and Ren, Jinke and Gao, Anningzhe and Guo, Dandan and Wan, Xiang and Zha, Hongyuan},
  year = {2025},
  month = apr,
  journal = {Transactions on Machine Learning Research},
  issn = {2835-8856},
  langid = {english}
}

@article{shiCLIPGuidedFederatedLearning2024,
  title = {{{CLIP-Guided Federated Learning}} on {{Heterogeneity}} and {{Long-Tailed Data}}},
  author = {Shi, Jiangming and Zheng, Shanshan and Yin, Xiangbo and Lu, Yang and Xie, Yuan and Qu, Yanyun},
  year = {2024},
  month = mar,
  journal = {Proceedings of the AAAI Conference on Artificial Intelligence},
  volume = {38},
  number = {13},
  pages = {14955--14963},
  issn = {2374-3468},
  doi = {10.1609/aaai.v38i13.29416},
  copyright = {Copyright (c) 2024 Association for the Advancement of Artificial Intelligence},
  langid = {english}
}

@article{gama2014survey,
  title={A survey on concept drift adaptation},
  author={Gama, Jo{\~a}o and {\v{Z}}liobait{\.e}, Indr{\.e} and Bifet, Albert and Pechenizkiy, Mykola and Bouchachia, Abdelhamid},
  journal={ACM computing surveys (CSUR)},
  volume={46},
  number={4},
  pages={1--37},
  year={2014},
  publisher={ACM New York, NY, USA}
}

@inproceedings{changUnifiedDomainGeneralization2024,
  title = {Unified {{Domain Generalization}} and {{Adaptation}} for {{Multi-View 3D Object Detection}}},
  booktitle = {Advances in {{Neural Information Processing Systems}}},
  author = {Chang, Gyusam and Lee, Jiwon and Kim, Donghyun and Kim, Jinkyu and Lee, Dongwook and Ji, Daehyun and Jang, Sujin and Kim, Sangpil},
  year = {2024},
  volume = {37},
  pages = {58498--58524},
  publisher = {Curran Associates, Inc.}
}

@inproceedings{chenMultiPromptAlignmentMultiSource2023,
  title = {Multi-{{Prompt Alignment}} for {{Multi-Source Unsupervised Domain Adaptation}}},
  booktitle = {Advances in {{Neural Information Processing Systems}}},
  author = {Chen, Haoran and Han, Xintong and Wu, Zuxuan and Jiang, Yu-Gang},
  year = {2023},
  volume = {36},
  pages = {74127--74139},
  publisher = {Curran Associates, Inc.}
}

@inproceedings{duDiffusionBasedProbabilisticUncertainty2023,
  title = {Diffusion-{{Based Probabilistic Uncertainty Estimation}} for {{Active Domain Adaptation}}},
  booktitle = {Advances in {{Neural Information Processing Systems}}},
  author = {Du, Zhekai and Li, Jingjing},
  year = {2023},
  volume = {36},
  pages = {17129--17155},
  publisher = {Curran Associates, Inc.}
}

@inproceedings{fengOpenCompoundDomain2023,
  title = {Open {{Compound Domain Adaptation}} with {{Object Style Compensation}} for {{Semantic Segmentation}}},
  booktitle = {Advances in {{Neural Information Processing Systems}}},
  author = {Feng, Tingliang and Shi, Hao and Liu, Xueyang and Feng, Wei and Wan, Liang and Zhou, Yanlin and Lin, Di},
  year = {2023},
  volume = {36},
  pages = {63136--63149},
  publisher = {Curran Associates, Inc.}
}

@inproceedings{heDomainAdaptationTime2023,
  title = {Domain {{Adaptation}} for {{Time Series Under Feature}} and {{Label Shifts}}},
  booktitle = {Proceedings of the 40th {{International Conference}} on {{Machine Learning}}},
  author = {He, Huan and Queen, Owen and Koker, Teddy and Cuevas, Consuelo and Tsiligkaridis, Theodoros and Zitnik, Marinka},
  year = {2023},
  month = jul,
  pages = {12746--12774},
  publisher = {PMLR},
  issn = {2640-3498},
  langid = {english}
}

@inproceedings{jiaDapperFLDomainAdaptive2024,
  title = {{{DapperFL}}: {{Domain Adaptive Federated Learning}} with {{Model Fusion Pruning}} for {{Edge Devices}}},
  shorttitle = {{{DapperFL}}},
  booktitle = {Advances in {{Neural Information Processing Systems}}},
  author = {Jia, Yongzhe and Zhang, Xuyun and Hu, Hongsheng and Choo, Kim-Kwang Raymond and Qi, Lianyong and Xu, Xiaolong and Beheshti, Amin and Dou, Wanchun},
  year = {2024},
  volume = {37},
  pages = {13099--13123},
  publisher = {Curran Associates, Inc.}
}

@inproceedings{liSubspaceIdentificationMultiSource2023,
  title = {Subspace {{Identification}} for {{Multi-Source Domain Adaptation}}},
  booktitle = {Advances in {{Neural Information Processing Systems}}},
  author = {Li, Zijian and Cai, Ruichu and Chen, Guangyi and Sun, Boyang and Hao, Zhifeng and Zhang, Kun},
  year = {2023},
  volume = {36},
  pages = {34504--34518},
  publisher = {Curran Associates, Inc.}
}

@inproceedings{liuBoostingTransferabilityDiscriminability2024,
  title = {Boosting {{Transferability}} and {{Discriminability}} for {{Time Series Domain Adaptation}}},
  booktitle = {Advances in {{Neural Information Processing Systems}}},
  author = {Liu, Mingyang and Chen, Xinyang and Shu, Yang and Li, Xiucheng and Guan, Weili and Nie, Liqiang},
  year = {2024},
  volume = {37},
  pages = {100402--100427},
  publisher = {Curran Associates, Inc.}
}

@inproceedings{luStyleAdaptationUncertainty2024,
  title = {Style {{Adaptation}} and {{Uncertainty Estimation}} for {{Multi-Source Blended-Target Domain Adaptation}}},
  booktitle = {Advances in {{Neural Information Processing Systems}}},
  author = {Lu, Yuwu and Huang, Haoyu and Hu, Xue},
  year = {2024},
  volume = {37},
  pages = {87042--87060},
  publisher = {Curran Associates, Inc.}
}

@inproceedings{saberiGradualDomainAdaptation2024,
  title = {Gradual {{Domain Adaptation}} via {{Manifold-Constrained Distributionally Robust Optimization}}},
  booktitle = {Advances in {{Neural Information Processing Systems}}},
  author = {Saberi, Amirhossein and Najafi, Amir and Behjati, Amin and Emrani, Ala and Zolfit, Yasaman and Shadrooy, Mahdi and Motahari, Abolfazl and Khalaj, Babak H.},
  year = {2024},
  volume = {37},
  pages = {73693--73725},
  publisher = {Curran Associates, Inc.}
}

@inproceedings{sunAdversarialAlignmentAnchor2025,
  title = {Adversarial {{Alignment}} with {{Anchor Dragging Drift}} ({{A}}{\textasciicircum}{{3D}}{\textasciicircum}2): {{Multimodal Domain Adaptation}} with {{Partially Shifted Modalities}}},
  shorttitle = {Adversarial {{Alignment}} with {{Anchor Dragging Drift}} ({{A}}{\textasciicircum}{{3D}}{\textasciicircum}2)},
  booktitle = {Proceedings of the 63rd {{Annual Meeting}} of the {{Association}} for {{Computational Linguistics}} ({{Volume}} 1: {{Long Papers}})},
  author = {Sun, Jun and Zhang, Xinxin and Hong, Simin and Zhu, Jian and Zeng, Lingfang},
  editor = {Che, Wanxiang and Nabende, Joyce and Shutova, Ekaterina and Pilehvar, Mohammad Taher},
  year = {2025},
  month = jul,
  pages = {19680--19690},
  publisher = {Association for Computational Linguistics},
  address = {Vienna, Austria},
  doi = {10.18653/v1/2025.acl-long.967},
  isbn = {979-8-89176-251-0}
}

@inproceedings{wangFDivergencePrincipledDomain2024,
  title = {On F-{{Divergence Principled Domain Adaptation}}: {{An Improved Framework}}},
  shorttitle = {On F-{{Divergence Principled Domain Adaptation}}},
  booktitle = {Advances in {{Neural Information Processing Systems}}},
  author = {Wang, Ziqiao and Mao, Yongyi},
  year = {2024},
  volume = {37},
  pages = {6711--6748},
  publisher = {Curran Associates, Inc.}
}

@inproceedings{weiUnsupervisedVideoDomain2023,
  title = {Unsupervised {{Video Domain Adaptation}} for {{Action Recognition}}: {{A Disentanglement Perspective}}},
  shorttitle = {Unsupervised {{Video Domain Adaptation}} for {{Action Recognition}}},
  booktitle = {Advances in {{Neural Information Processing Systems}}},
  author = {Wei, Pengfei and Kong, Lingdong and Qu, Xinghua and Ren, Yi and Xu, Zhiqiang and Jiang, Jing and Yin, Xiang},
  year = {2023},
  volume = {36},
  pages = {17623--17642},
  publisher = {Curran Associates, Inc.}
}

@inproceedings{xiaoSPAGraphSpectral2023,
  title = {{{SPA}}: {{A Graph Spectral Alignment Perspective}} for {{Domain Adaptation}}},
  shorttitle = {{{SPA}}},
  booktitle = {Advances in {{Neural Information Processing Systems}}},
  author = {Xiao, Zhiqing and Wang, Haobo and Jin, Ying and Feng, Lei and Chen, Gang and Huang, Fei and Zhao, Junbo},
  year = {2023},
  volume = {36},
  pages = {37252--37272},
  publisher = {Curran Associates, Inc.}
}

@article{faridTaskDrivenDetectionDistribution2025,
  title = {Task-{{Driven Detection}} of {{Distribution Shifts With Statistical Guarantees}} for {{Robot Learning}}},
  author = {Farid, Alec and Veer, Sushant and Pachisia, Divyanshu and Majumdar, Anirudha},
  year = {2025},
  journal = {IEEE Transactions on Robotics},
  volume = {41},
  pages = {926--945},
  issn = {1941-0468},
  doi = {10.1109/TRO.2024.3521963}
}

@article{yuRuntoRunControlChemical2020,
  title = {Run-to-{{Run Control}} of {{Chemical Mechanical Polishing Process Based}} on {{Deep Reinforcement Learning}}},
  author = {Yu, Jianbo and Guo, Peng},
  year = 2020,
  month = aug,
  journal = {IEEE Transactions on Semiconductor Manufacturing},
  volume = {33},
  number = {3},
  pages = {454--465},
  issn = {1558-2345},
  doi = {10.1109/TSM.2020.3002896},
  urldate = {2025-10-10}
}

@article{shayestehAutomatedConceptDrift2022a,
  title = {Automated {{Concept Drift Handling}} for {{Fault Prediction}} in {{Edge Clouds Using Reinforcement Learning}}},
  author = {Shayesteh, Behshid and Fu, Chunyan and Ebrahimzadeh, Amin and Glitho, Roch H.},
  year = 2022,
  month = jun,
  journal = {IEEE Transactions on Network and Service Management},
  volume = {19},
  number = {2},
  pages = {1321--1335},
  issn = {1932-4537},
  doi = {10.1109/TNSM.2022.3153279},
  urldate = {2025-10-10}
}

@inproceedings{kimSufficientInvariantLearning2025,
  title = {Sufficient {{Invariant Learning}} for {{Distribution Shift}}},
  booktitle = {Proceedings of the {{IEEE}}/{{CVF Conference}} on {{Computer Vision}} and {{Pattern Recognition}}},
  author = {Kim, Taero and Park, Subeen and Lim, Sungjun and Jung, Yonghan and Muandet, Krikamol and Song, Kyungwoo},
  year = {2025},
  pages = {4958--4967},
  langid = {english}
}

@article{kimTestTimeAdaptationInduces2024,
  title = {Test-{{Time Adaptation Induces Stronger Accuracy}} and {{Agreement-on-the-Line}}},
  author = {Kim, Eungyeup and Sun, Mingjie and Baek, Christina and Raghunathan, Aditi and Kolter, J. Zico},
  year = {2024},
  month = dec,
  journal = {Advances in Neural Information Processing Systems},
  volume = {37},
  pages = {120184--120220},
  langid = {english}
}

@article{liaoFOOGDFederatedCollaboration2024,
  title = {{{FOOGD}}: {{Federated Collaboration}} for {{Both Out-of-distribution Generalization}} and {{Detection}}},
  shorttitle = {{{FOOGD}}},
  author = {Liao, Xinting and Liu, Weiming and Zhou, Pengyang and Yu, Fengyuan and Xu, Jiahe and Wang, Jun and Wang, Wenjie and Chen, Chaochao and Zheng, Xiaolin},
  year = {2024},
  month = dec,
  journal = {Advances in Neural Information Processing Systems},
  volume = {37},
  pages = {132908--132945},
  langid = {english}
}

@inproceedings{liGraphStructureExtrapolation2024,
  title = {Graph {{Structure Extrapolation}} for {{Out-of-Distribution Generalization}}},
  booktitle = {Forty-First {{International Conference}} on {{Machine Learning}}},
  author = {Li, Xiner and Gui, Shurui and Luo, Youzhi and Ji, Shuiwang},
  year = {2024},
  month = jun,
  langid = {english}
}

@article{liLetInvariantLearning2025,
  title = {Let {{Invariant Learning Inspire Neighbor-shift Generalization}} on {{Graphs}}},
  author = {Li, Jiaxing and Gao, Jiayi and Gu, Binhao and Kong, Youyong},
  year = {2025},
  journal = {IEEE Transactions on Artificial Intelligence},
  pages = {1--12},
  issn = {2691-4581},
  doi = {10.1109/TAI.2025.3605894}
}

@inproceedings{liuTimeseriesForecastingOutofdistribution2024,
  title = {Time-Series Forecasting for out-of-Distribution Generalization Using Invariant Learning},
  booktitle = {Proceedings of the 41st {{International Conference}} on {{Machine Learning}}},
  author = {Liu, Haoxin and Kamarthi, Harshavardhan and Kong, Lingkai and Zhao, Zhiyuan and Zhang, Chao and Prakash, B. Aditya},
  year = {2024},
  month = jul,
  series = {{{ICML}}'24},
  volume = {235},
  pages = {31312--31325},
  publisher = {JMLR.org},
  address = {Vienna, Austria}
}

@article{liuUDDATCUnsupervisedRealTime2025,
  title = {{{UDDA-TC}}: {{Unsupervised Real-Time Drift Detection}} and {{Adaptation}} for {{Continual Traffic Classification}} in {{Mobile Edge Computing}}},
  shorttitle = {{{UDDA-TC}}},
  author = {Liu, Minyao and Wang, Pan and Li, Zeyi and Ye, Yingchun and Wang, Zixuan and Chen, Xuejiao},
  year = {2025},
  journal = {IEEE Transactions on Consumer Electronics},
  pages = {1--1},
  issn = {1558-4127},
  doi = {10.1109/TCE.2025.3579882}
}

@article{neoMaxEntLossConstrained2024,
  title = {{{MaxEnt Loss}}: {{Constrained Maximum Entropy}} for {{Calibration}} under {{Out-of-Distribution Shift}}},
  shorttitle = {{{MaxEnt Loss}}},
  author = {Neo, Dexter and Winkler, Stefan and Chen, Tsuhan},
  year = {2024},
  month = mar,
  journal = {Proceedings of the AAAI Conference on Artificial Intelligence},
  volume = {38},
  number = {19},
  pages = {21463--21472},
  issn = {2374-3468},
  doi = {10.1609/aaai.v38i19.30143},
  copyright = {Copyright (c) 2024 Association for the Advancement of Artificial Intelligence},
  langid = {english}
}

@inproceedings{shenOptimizingOODDetection2024,
  title = {Optimizing {{OOD Detection}} in {{Molecular Graphs}}: {{A Novel Approach}} with {{Diffusion Models}}},
  shorttitle = {Optimizing {{OOD Detection}} in {{Molecular Graphs}}},
  booktitle = {Proceedings of the 30th {{ACM SIGKDD Conference}} on {{Knowledge Discovery}} and {{Data Mining}}},
  author = {Shen, Xu and Wang, Yili and Zhou, Kaixiong and Pan, Shirui and Wang, Xin},
  year = {2024},
  month = aug,
  series = {{{KDD}} '24},
  pages = {2640--2650},
  publisher = {Association for Computing Machinery},
  address = {New York, NY, USA},
  doi = {10.1145/3637528.3671785},
  isbn = {979-8-4007-0490-1}
}

@article{yangDualTestTimeTraining2025,
  title = {Dual {{Test-Time Training}} for {{Out-of-Distribution Recommender System}}},
  author = {Yang, Xihong and Wang, Yiqi and Chen, Jin and Fan, Wenqi and Zhao, Xiangyu and Zhu, En and Liu, Xinwang and Lian, Defu},
  year = {2025},
  month = jun,
  journal = {IEEE Transactions on Knowledge and Data Engineering},
  volume = {37},
  number = {6},
  pages = {3312--3326},
  issn = {1558-2191},
  doi = {10.1109/TKDE.2025.3548160}
}

@inproceedings{yangOODDTesttimeOutofDistribution2025,
  title = {{{OODD}}: {{Test-time Out-of-Distribution Detection}} with {{Dynamic Dictionary}}},
  shorttitle = {{{OODD}}},
  booktitle = {Proceedings of the {{IEEE}}/{{CVF Conference}} on {{Computer Vision}} and {{Pattern Recognition}}},
  author = {Yang, Yifeng and Zhu, Lin and Sun, Zewen and Liu, Hengyu and Gu, Qinying and Ye, Nanyang},
  year = {2025},
  pages = {30630--30639},
  langid = {english}
}

@inproceedings{zhaoProactiveModelAdaptation2025,
  title = {Proactive {{Model Adaptation Against Concept Drift}} for {{Online Time Series Forecasting}}},
  booktitle = {Proceedings of the 31st {{ACM SIGKDD Conference}} on {{Knowledge Discovery}} and {{Data Mining V}}.1},
  author = {Zhao, Lifan and Shen, Yanyan},
  year = {2025},
  month = jul,
  series = {{{KDD}} '25},
  pages = {2020--2031},
  publisher = {Association for Computing Machinery},
  address = {New York, NY, USA},
  doi = {10.1145/3690624.3709210},
  isbn = {979-8-4007-1245-6}
}

@article{zouGeSSBenchmarkingGeometric2024,
  title = {{{GeSS}}: {{Benchmarking Geometric Deep Learning}} under {{Scientific Applications}} with {{Distribution Shifts}}},
  shorttitle = {{{GeSS}}},
  author = {Zou, Deyu and Liu, Shikun and Miao, Siqi and Fung, Victor and Chang, Shiyu and Li, Pan},
  year = {2024},
  month = dec,
  journal = {Advances in Neural Information Processing Systems},
  volume = {37},
  pages = {92499--92528},
  langid = {english}
}

@inproceedings{aguiarEnhancingConceptDrift2023,
  title = {Enhancing {{Concept Drift Detection}} in {{Drifting}} and {{Imbalanced Data Streams}} through {{Meta-Learning}}},
  booktitle = {2023 {{IEEE International Conference}} on {{Big Data}} ({{BigData}})},
  author = {Aguiar, Gabriel Jonas and Cano, Alberto},
  year = {2023},
  month = dec,
  pages = {2648--2657},
  doi = {10.1109/BigData59044.2023.10386364}
}

@article{mawuliFedStreamPrototypeBasedFederated2023,
  title = {{{FedStream}}: {{Prototype-Based Federated Learning}} on {{Distributed Concept-Drifting Data Streams}}},
  shorttitle = {{{FedStream}}},
  author = {Mawuli, Cobbinah B. and Che, Liwei and Kumar, Jay and Din, Salah Ud and Qin, Zhili and Yang, Qinli and Shao, Junming},
  year = {2023},
  month = nov,
  journal = {IEEE Transactions on Systems, Man, and Cybernetics: Systems},
  volume = {53},
  number = {11},
  pages = {7112--7124},
  issn = {2168-2232},
  doi = {10.1109/TSMC.2023.3293462}
}

@article{mehrtensBenchmarkingCommonUncertainty2023,
  title = {Benchmarking Common Uncertainty Estimation Methods with Histopathological Images under Domain Shift and Label Noise},
  author = {Mehrtens, Hendrik A. and Kurz, Alexander and Bucher, Tabea-Clara and Brinker, Titus J.},
  year = {2023},
  month = oct,
  journal = {Medical Image Analysis},
  volume = {89},
  pages = {102914},
  issn = {1361-8415},
  doi = {10.1016/j.media.2023.102914}
}

@inproceedings{naLabelNoiseRobustDiffusion2023,
  title = {Label-{{Noise Robust Diffusion Models}}},
  booktitle = {The {{Twelfth International Conference}} on {{Learning Representations}}},
  author = {Na, Byeonghu and Kim, Yeongmin and Bae, HeeSun and Lee, Jung Hyun and Kwon, Se Jung and Kang, Wanmo and Moon, Il-chul},
  year = {2023},
  month = oct,
  langid = {english}
}

@inproceedings{yangReCDAConceptDrift2024,
  title = {{{ReCDA}}: {{Concept Drift Adaptation}} with {{Representation Enhancement}} for {{Network Intrusion Detection}}},
  shorttitle = {{{ReCDA}}},
  booktitle = {Proceedings of the 30th {{ACM SIGKDD Conference}} on {{Knowledge Discovery}} and {{Data Mining}}},
  author = {Yang, Shuo and Zheng, Xinran and Li, Jinze and Xu, Jinfeng and Wang, Xingjun and Ngai, Edith C. H.},
  year = {2024},
  month = aug,
  series = {{{KDD}} '24},
  pages = {3818--3828},
  publisher = {Association for Computing Machinery},
  address = {New York, NY, USA},
  doi = {10.1145/3637528.3672007},
  isbn = {979-8-4007-0490-1}
}

@article{yuTreatNoiseDomain2023,
  title = {Treat {{Noise}} as {{Domain Shift}}: {{Noise Feature Disentanglement}} for {{Underwater Perception}} and {{Maritime Surveys}} in {{Side-Scan Sonar Images}}},
  shorttitle = {Treat {{Noise}} as {{Domain Shift}}},
  author = {Yu, Yongcan and Zhao, Jianhu and Huang, Chao and Zhao, Xi},
  year = {2023},
  journal = {IEEE Transactions on Geoscience and Remote Sensing},
  volume = {61},
  pages = {1--15},
  issn = {1558-0644},
  doi = {10.1109/TGRS.2023.3322787}
}

@article{zhaoFedFMFederatedFewshot2025,
  title = {{{FedFM}}: {{A}} Federated Few-Shot Learning Method by Comparison Network and Model Calibration},
  shorttitle = {{{FedFM}}},
  author = {Zhao, Chen and Bao, Shudi and Chen, Meng and Gao, Zhipeng and Xiao, Kaile and Dai, Peng},
  year = {2025},
  month = jan,
  journal = {Knowledge-Based Systems},
  volume = {309},
  pages = {112848},
  issn = {0950-7051},
  doi = {10.1016/j.knosys.2024.112848}
}

@article{ajayDistributionallyAdaptiveMeta2022,
  title = {Distributionally {{Adaptive Meta Reinforcement Learning}}},
  author = {Ajay, Anurag and Gupta, Abhishek and Ghosh, Dibya and Levine, Sergey and Agrawal, Pulkit},
  year = 2022,
  month = dec,
  journal = {Advances in Neural Information Processing Systems},
  volume = {35},
  pages = {25856--25869},
  langid = {english}
}

@article{caoDriftShieldAutonomousFraud2025,
  title = {{{DriftShield}}: {{Autonomous Fraud Detection}} via {{Actor-Critic Reinforcement Learning With Dynamic Feature Reweighting}}},
  shorttitle = {{{DriftShield}}},
  author = {Cao, Jialei and Zheng, Wenxia and Ge, Yao and Wang, Jiyuan},
  year = 2025,
  journal = {IEEE Open Journal of the Computer Society},
  volume = {6},
  pages = {1166--1177},
  issn = {2644-1268},
  doi = {10.1109/OJCS.2025.3587001}
}

@article{gangulyOnlineFederatedLearning2024,
  title = {Online {{Federated Learning}} via {{Non-Stationary Detection}} and {{Adaptation Amidst Concept Drift}}},
  author = {Ganguly, Bhargav and Aggarwal, Vaneet},
  year = 2024,
  month = feb,
  journal = {IEEE/ACM Transactions on Networking},
  volume = {32},
  number = {1},
  pages = {643--653},
  issn = {1558-2566},
  doi = {10.1109/TNET.2023.3294366}
}

@article{lianGOFLAccurateEfficient2024,
  title = {{{GOFL}}: {{An Accurate}} and {{Efficient Federated Learning Framework Based}} on {{Gradient Optimization}} in {{Heterogeneous IoT Systems}}},
  shorttitle = {{{GOFL}}},
  author = {Lian, Zirui and Cao, Jing and Zhu, Zongwei and Zhou, Xuehai and Liu, Weihong},
  year = {2024},
  month = apr,
  journal = {IEEE Internet of Things Journal},
  volume = {11},
  number = {7},
  pages = {12459--12474},
  issn = {2327-4662},
  doi = {10.1109/JIOT.2023.3333419}
}

@article{maniasModelDriftDynamic2023,
  title = {Model {{Drift}} in {{Dynamic Networks}}},
  author = {Manias, Dimitrios Michael and Chouman, Ali and Shami, Abdallah},
  year = 2023,
  month = oct,
  journal = {IEEE Communications Magazine},
  volume = {61},
  number = {10},
  pages = {78--84},
  issn = {1558-1896},
  doi = {10.1109/MCOM.003.2200306}
}

@inproceedings{perdomoPerformativePrediction2020,
  title = {Performative {{Prediction}}},
  booktitle = {Proceedings of the 37th {{International Conference}} on {{Machine Learning}}},
  author = {Perdomo, Juan and Zrnic, Tijana and {Mendler-D{\"u}nner}, Celestine and Hardt, Moritz},
  year = 2020,
  month = nov,
  pages = {7599--7609},
  publisher = {PMLR},
  issn = {2640-3498},
  langid = {english}
}

@article{poddubnyyOnlineEVCharging2023,
  title = {Online {{EV}} Charging Controlled by Reinforcement Learning with Experience Replay},
  author = {Poddubnyy, Andrey and Nguyen, Phuong and Slootweg, Han},
  year = 2023,
  month = dec,
  journal = {Sustainable Energy, Grids and Networks},
  volume = {36},
  pages = {101162},
  issn = {2352-4677},
  doi = {10.1016/j.segan.2023.101162}
}

@inproceedings{schweighoferDatasetPerspectiveOffline2022,
  title = {A {{Dataset Perspective}} on {{Offline Reinforcement Learning}}},
  booktitle = {Proceedings of {{The}} 1st {{Conference}} on {{Lifelong Learning Agents}}},
  author = {Schweighofer, Kajetan and Dinu, Marius-constantin and Radler, Andreas and Hofmarcher, Markus and Patil, Vihang Prakash and {Bitto-nemling}, Angela and {Eghbal-zadeh}, Hamid and Hochreiter, Sepp},
  year = 2022,
  month = nov,
  pages = {470--517},
  publisher = {PMLR},
  issn = {2640-3498},
  langid = {english}
}

@inproceedings{steinReflectiveLearningClassifier2021,
  title = {Reflective {{Learning Classifier Systems}} for {{Self-Adaptive}} and {{Self-Organising Agents}}},
  booktitle = {2021 {{IEEE International Conference}} on {{Autonomic Computing}} and {{Self-Organizing Systems Companion}} ({{ACSOS-C}})},
  author = {Stein, Anthony and Tomforde, Sven},
  year = 2021,
  month = sep,
  pages = {139--145},
  doi = {10.1109/ACSOS-C52956.2021.00043}
}

@article{wuGeneralizedRISTile2024,
  title = {Generalized {{RIS Tile Exclusion Strategy}} for {{Indoor mmWave Channels Under Concept Drift}}},
  author = {Wu, Zi-Yang and Ismail, Muhammad},
  year = 2024,
  month = oct,
  journal = {IEEE Transactions on Wireless Communications},
  volume = {23},
  number = {10},
  pages = {13484--13498},
  issn = {1558-2248},
  doi = {10.1109/TWC.2024.3402267}
}

@article{zhangOneNetEnhancingTime2023,
  title = {{{OneNet}}: {{Enhancing Time Series Forecasting Models}} under {{Concept Drift}} by {{Online Ensembling}}},
  shorttitle = {{{OneNet}}},
  author = {Zhang, Yifan and Wen, Qingsong and Wang, Xue and Chen, Weiqi and Sun, Liang and Zhang, Zhang and Wang, Liang and Jin, Rong and Tan, Tieniu},
  year = 2023,
  month = dec,
  journal = {Advances in Neural Information Processing Systems},
  volume = {36},
  pages = {69949--69980},
  langid = {english}
}

@inproceedings{zhaoPerformativeTimeSeriesForecasting2025,
  title = {Performative {{Time-Series Forecasting}}},
  booktitle = {Proceedings of the 31st {{ACM SIGKDD Conference}} on {{Knowledge Discovery}} and {{Data Mining V}}.2},
  author = {Zhao, Zhiyuan and Liu, Haoxin and Rodr{\'i}guez, Alexander and Prakash, B. Aditya},
  year = 2025,
  month = aug,
  series = {{{KDD}} '25},
  pages = {3968--3979},
  publisher = {Association for Computing Machinery},
  address = {New York, NY, USA},
  doi = {10.1145/3711896.3737078},
  isbn = {979-8-4007-1454-2}
}

@inproceedings{arteltUnsupervisedUnlearningConcept2023,
  title = {Unsupervised {{Unlearning}} of {{Concept Drift}} with {{Autoencoders}}},
  booktitle = {2023 {{IEEE Symposium Series}} on {{Computational Intelligence}} ({{SSCI}})},
  author = {Artelt, Andr{\'e} and Malialis, Kleanthis and Panayiotou, Christos G. and Polycarpou, Marios M. and Hammer, Barbara},
  year = 2023,
  month = dec,
  pages = {703--710},
  issn = {2472-8322},
  doi = {10.1109/SSCI52147.2023.10372001},
  langid = {american}
}

@article{chaiMalFSCILFewShotClassIncremental2025,
  title = {{{MalFSCIL}}: {{A Few-Shot Class-Incremental Learning Approach}} for {{Malware Detection}}},
  shorttitle = {{{MalFSCIL}}},
  author = {Chai, Yuhan and Chen, Ximing and Qiu, Jing and Du, Lei and Xiao, Yanjun and Feng, Qiying and Ji, Shouling and Tian, Zhihong},
  year = 2025,
  journal = {IEEE Transactions on Information Forensics and Security},
  volume = {20},
  pages = {2999--3014},
  issn = {1556-6021},
  doi = {10.1109/TIFS.2024.3516565}
}

@inproceedings{chenDynamicResidualClassifier2023,
  title = {Dynamic {{Residual Classifier}} for {{Class Incremental Learning}}},
  booktitle = {Proceedings of the {{IEEE}}/{{CVF International Conference}} on {{Computer Vision}}},
  author = {Chen, Xiuwei and Chang, Xiaobin},
  year = 2023,
  pages = {18743--18752},
  langid = {english}
}

@article{dengCentroidGuidedDomainIncremental2024,
  title = {Centroid-{{Guided Domain Incremental Learning}} for {{EEG-Based Seizure Prediction}}},
  author = {Deng, Zhiwei and Li, Chang and Song, Rencheng and Liu, Xiang and Qian, Ruobing and Chen, Xun},
  year = 2024,
  journal = {IEEE Transactions on Instrumentation and Measurement},
  volume = {73},
  pages = {1--13},
  issn = {1557-9662},
  doi = {10.1109/TIM.2023.3334330}
}

@inproceedings{heDYSONDynamicFeature2024,
  title = {{{DYSON}}: {{Dynamic Feature Space Self-Organization}} for {{Online Task-Free Class Incremental Learning}}},
  shorttitle = {{{DYSON}}},
  booktitle = {Proceedings of the {{IEEE}}/{{CVF Conference}} on {{Computer Vision}} and {{Pattern Recognition}}},
  author = {He, Yuhang and Chen, Yingjie and Jin, Yuhan and Dong, Songlin and Wei, Xing and Gong, Yihong},
  year = 2024,
  pages = {23741--23751},
  langid = {english}
}

@article{liConceptDriftAdaptation2024a,
  title = {Concept {{Drift Adaptation}} by {{Exploiting Drift Type}}},
  author = {Li, Jinpeng and Yu, Hang and Zhang, Zhenyu and Luo, Xiangfeng and Xie, Shaorong},
  year = 2024,
  month = feb,
  journal = {ACM Trans. Knowl. Discov. Data},
  volume = {18},
  number = {4},
  pages = {96:1--96:22},
  issn = {1556-4681},
  doi = {10.1145/3638777}
}

@inproceedings{liDynamicIntegrationTaskSpecific2025a,
  title = {Dynamic {{Integration}} of {{Task-Specific Adapters}} for {{Class Incremental Learning}}},
  booktitle = {Proceedings of the {{IEEE}}/{{CVF Conference}} on {{Computer Vision}} and {{Pattern Recognition}}},
  author = {Li, Jiashuo and Wang, Shaokun and Qian, Bo and He, Yuhang and Wei, Xing and Wang, Qiang and Gong, Yihong},
  year = 2025,
  pages = {30545--30555},
  langid = {english}
}

@inproceedings{panchalFlashConceptDrift2023b,
  title = {Flash: {{Concept Drift Adaptation}} in {{Federated Learning}}},
  shorttitle = {Flash},
  booktitle = {Proceedings of the 40th {{International Conference}} on {{Machine Learning}}},
  author = {Panchal, Kunjal and Choudhary, Sunav and Mitra, Subrata and Mukherjee, Koyel and Sarkhel, Somdeb and Mitra, Saayan and Guan, Hui},
  year = 2023,
  month = jul,
  pages = {26931--26962},
  publisher = {PMLR},
  issn = {2640-3498},
  langid = {english}
}

@article{pengUnsupervisedAdaptiveFleet2024a,
  title = {Unsupervised {{Adaptive Fleet Battery Pack Fault Detection With Concept Drift Under Evolving Environment}}},
  author = {Peng, Xiaomeng and Duan, Shiming and Sankavaram, Chaitanya and Jin, Xiaoning},
  year = 2024,
  month = jul,
  journal = {IEEE Transactions on Automation Science and Engineering},
  volume = {21},
  number = {3},
  pages = {2276--2288},
  issn = {1558-3783},
  doi = {10.1109/TASE.2024.3363002}
}

@inproceedings{petitFeTrILFeatureTranslation2023,
  title = {{{FeTrIL}}: {{Feature Translation}} for {{Exemplar-Free Class-Incremental Learning}}},
  shorttitle = {{{FeTrIL}}},
  booktitle = {Proceedings of the {{IEEE}}/{{CVF Winter Conference}} on {{Applications}} of {{Computer Vision}}},
  author = {Petit, Gr{\'e}goire and Popescu, Adrian and Schindler, Hugo and Picard, David and Delezoide, Bertrand},
  year = 2023,
  pages = {3911--3920},
  langid = {english}
}

@article{sunAntiforgettingIncrementalLearning2024,
  title = {Antiforgetting {{Incremental Learning Algorithm}} for {{Interval Type-2 Fuzzy Neural Network}}},
  author = {Sun, Chenxuan and Han, Honggui and Wu, Xiaolong and Yang, Hongyan},
  year = 2024,
  month = apr,
  journal = {IEEE Transactions on Fuzzy Systems},
  volume = {32},
  number = {4},
  pages = {1938--1950},
  issn = {1941-0034},
  doi = {10.1109/TFUZZ.2023.3336325}
}

@article{sunClassIncrementalLearning2023,
  title = {Class {{Incremental Learning}} Based on {{Identically Distributed Parallel One-Class Classifiers}}},
  author = {Sun, Wenju and Li, Qingyong and Zhang, Jing and Wang, Wen and Geng, YangLi-ao},
  year = 2023,
  month = nov,
  journal = {Neurocomputing},
  volume = {556},
  pages = {126579},
  issn = {0925-2312},
  doi = {10.1016/j.neucom.2023.126579}
}

@article{sunMOSModelSurgery2025,
  title = {{{MOS}}: {{Model Surgery}} for {{Pre-Trained Model-Based Class-Incremental Learning}}},
  shorttitle = {{{MOS}}},
  author = {Sun, Hai-Long and Zhou, Da-Wei and Zhao, Hanbin and Gan, Le and Zhan, De-Chuan and Ye, Han-Jia},
  year = 2025,
  month = apr,
  journal = {Proceedings of the AAAI Conference on Artificial Intelligence},
  volume = {39},
  number = {19},
  pages = {20699--20707},
  issn = {2374-3468},
  doi = {10.1609/aaai.v39i19.34281},
  copyright = {Copyright (c) 2025 Association for the Advancement of Artificial Intelligence},
  langid = {english}
}

@article{yuGeneralizedIncrementalLearning2025a,
author = {Yu, En and Lu, Jie and Zhang, Guangquan},
title = {Generalized Incremental Learning under Concept Drift across Evolving Data Streams},
year = {2026},
journal = {Proceedings of the ACM Web Conference 2026},
pages = {3905–3916},
numpages = {12},
series = {WWW '26},
doi = {10.1145/3774904.3792379}
}

@article{yuOnlineBoostingAdaptive2024a,
  title = {Online {{Boosting Adaptive Learning}} under {{Concept Drift}} for {{Multistream Classification}}},
  author = {Yu, En and Lu, Jie and Zhang, Bin and Zhang, Guangquan},
  year = 2024,
  month = mar,
  journal = {Proceedings of the AAAI Conference on Artificial Intelligence},
  volume = {38},
  number = {15},
  pages = {16522--16530},
  issn = {2374-3468},
  doi = {10.1609/aaai.v38i15.29590},
  copyright = {Copyright (c) 2024 Association for the Advancement of Artificial Intelligence},
  langid = {english}
}

@article{zhangOneNetEnhancingTime2023a,
  title = {{{OneNet}}: {{Enhancing Time Series Forecasting Models}} under {{Concept Drift}} by {{Online Ensembling}}},
  shorttitle = {{{OneNet}}},
  author = {Zhang, Yifan and Wen, Qingsong and Wang, Xue and Chen, Weiqi and Sun, Liang and Zhang, Zhang and Wang, Liang and Jin, Rong and Tan, Tieniu},
  year = 2023,
  month = dec,
  journal = {Advances in Neural Information Processing Systems},
  volume = {36},
  pages = {69949--69980},
  langid = {english}
}

@article{zhouIndustrialFaultDiagnosis2025,
  title = {Industrial {{Fault Diagnosis With Incremental Learning Capability Under Varying Sensory Data}}},
  author = {Zhou, Han and Yin, Hongpeng and Qin, Yan and Yuen, Chau},
  year = 2025,
  month = feb,
  journal = {IEEE Transactions on Systems, Man, and Cybernetics: Systems},
  volume = {55},
  number = {2},
  pages = {1322--1333},
  issn = {2168-2232},
  doi = {10.1109/TSMC.2024.3500019}
}

@article{bayramConceptDriftModel2022a,
  title = {From Concept Drift to Model Degradation: {{An}} Overview on Performance-Aware Drift Detectors},
  shorttitle = {From Concept Drift to Model Degradation},
  author = {Bayram, Firas and Ahmed, Bestoun S. and Kassler, Andreas},
  year = 2022,
  month = jun,
  journal = {Knowledge-Based Systems},
  volume = {245},
  pages = {108632},
  issn = {0950-7051},
  doi = {10.1016/j.knosys.2022.108632}
}

@article{hanSurveyActivePassive2022,
  title = {A Survey of Active and Passive Concept Drift Handling Methods},
  author = {Han, Meng and Chen, Zhiqiang and Li, Muhang and Wu, Hongxin and Zhang, Xilong},
  year = 2022,
  journal = {Computational Intelligence},
  volume = {38},
  number = {4},
  pages = {1492--1535},
  issn = {1467-8640},
  doi = {10.1111/coin.12520},
  copyright = {{\copyright} 2022 Wiley Periodicals LLC.},
  langid = {english}
}

@article{limaLearningConceptDrift2022,
  title = {Learning {{Under Concept Drift}} for {{Regression}}---{{A Systematic Literature Review}}},
  author = {Lima, Mar{\'i}lia and Neto, Manoel and Filho, Telmo Silva and {de A. Fagundes}, Roberta A.},
  year = 2022,
  journal = {IEEE Access},
  volume = {10},
  pages = {45410--45429},
  issn = {2169-3536},
  doi = {10.1109/ACCESS.2022.3169785}
}

@article{luLearningConceptDrift2019a,
  title = {Learning under {{Concept Drift}}: {{A Review}}},
  shorttitle = {Learning under {{Concept Drift}}},
  author = {Lu, Jie and Liu, Anjin and Dong, Fan and Gu, Feng and Gama, Jo{\~a}o and Zhang, Guangquan},
  year = 2019,
  month = dec,
  journal = {IEEE Transactions on Knowledge and Data Engineering},
  volume = {31},
  number = {12},
  pages = {2346--2363},
  issn = {1558-2191},
  doi = {10.1109/TKDE.2018.2876857}
}

@article{satoSurveyConceptDrift2021a,
  title = {A {{Survey}} on {{Concept Drift}} in {{Process Mining}}},
  author = {Sato, Denise Maria Vecino and De Freitas, Sheila Cristiana and Barddal, Jean Paul and Scalabrin, Edson Emilio},
  year = 2021,
  month = oct,
  journal = {ACM Comput. Surv.},
  volume = {54},
  number = {9},
  pages = {189:1--189:38},
  issn = {0360-0300},
  doi = {10.1145/3472752}
}

@inproceedings{babendererdeFederatedContinualDynamicSegmentation2025,
  title = {Federated-{{Continual Dynamic Segmentation}} of {{Histopathology Guided}} by {{Barlow Continuity}}},
  booktitle = {2025 {{IEEE}}/{{CVF Winter Conference}} on {{Applications}} of {{Computer Vision}} ({{WACV}})},
  author = {Babendererde, Niklas and Zhu, Haozhe and Fuchs, Moritz and Stieber, Jonathan and Mukhopadhyay, Anirban},
  year = {2025},
  month = feb,
  pages = {3752--3761},
  issn = {2642-9381},
  doi = {10.1109/WACV61041.2025.00369}
}

@inproceedings{kumariContinualDomainIncremental2024,
  title = {Continual {{Domain Incremental Learning}} for~{{Privacy-Aware Digital Pathology}}},
  booktitle = {Medical {{Image Computing}} and {{Computer Assisted Intervention}} -- {{MICCAI}} 2024},
  author = {Kumari, Pratibha and Reisenb{\"u}chler, Daniel and Luttner, Lucas and Schaadt, Nadine S. and Feuerhake, Friedrich and Merhof, Dorit},
  editor = {Linguraru, Marius George and Dou, Qi and Feragen, Aasa and Giannarou, Stamatia and Glocker, Ben and Lekadir, Karim and Schnabel, Julia A.},
  year = {2024},
  pages = {34--44},
  publisher = {Springer Nature Switzerland},
  address = {Cham},
  doi = {10.1007/978-3-031-72390-2_4},
  isbn = {978-3-031-72390-2},
  langid = {english}
}

@inproceedings{rahmanDecouplingClinicalClassAgnostic2026,
  title = {Decoupling {{Clinical}} and~{{Class-Agnostic Features}} for~{{Reliable Few-Shot Adaptation Under Shift}}},
  booktitle = {Uncertainty for {{Safe Utilization}} of {{Machine Learning}} in {{Medical Imaging}}},
  author = {Rahman, Umaima and Imam, Raza and Yaqub, Mohammad and Mahapatra, Dwarikanath},
  editor = {Sudre, Carole H. and Hoque, Mobarak I. and Mehta, Raghav and Ouyang, Cheng and Qin, Chen and Rakic, Marianne and Wells, William M.},
  year = {2026},
  pages = {123--133},
  publisher = {Springer Nature Switzerland},
  address = {Cham},
  doi = {10.1007/978-3-032-06593-3_12},
  isbn = {978-3-032-06593-3},
  langid = {english}
}

@inproceedings{saileClientSideAdaptationConcept2024,
  title = {Client-{{Side Adaptation}} to {{Concept Drift}} in {{Federated Learning}}},
  booktitle = {2024 2nd {{International Conference}} on {{Federated Learning Technologies}} and {{Applications}} ({{FLTA}})},
  author = {Saile, Finn and Thomas, Julius and Kaaser, Dominik and Schulte, Stefan},
  year = {2024},
  month = sep,
  pages = {71--78},
  doi = {10.1109/FLTA63145.2024.10840058}
}

@inproceedings{thomasAdaptionSelectionClient2025a,
  title = {Adaption via~{{Selection}}: {{On Client Selection}} to~{{Counter Concept Drift}} in~{{Federated Learning}}},
  shorttitle = {Adaption via~{{Selection}}},
  booktitle = {Service-{{Oriented}} and {{Cloud Computing}}},
  author = {Thomas, Julius and Saile, Finn and Fischer, Mathias and Kaaser, Dominik and Schulte, Stefan},
  editor = {Pahl, Claus and Janes, Andrea and Cerny, Tomas and Lenarduzzi, Valentina and Esposito, Matteo},
  year = {2025},
  pages = {3--17},
  publisher = {Springer Nature Switzerland},
  address = {Cham},
  doi = {10.1007/978-3-031-84617-5_1},
  isbn = {978-3-031-84617-5},
  langid = {english}
}

@article{wangFedSiMSimilarityMetric2023,
  title = {{{FedSiM}}: A Similarity Metric Federal Learning Mechanism Based on Stimulus Response Method with {{Non-IID}} Data},
  shorttitle = {{{FedSiM}}},
  author = {Wang, Shuangzhong and Zhang, Ying},
  year = {2023},
  month = sep,
  journal = {Measurement Science and Technology},
  volume = {34},
  number = {12},
  pages = {125045},
  publisher = {IOP Publishing},
  issn = {0957-0233},
  doi = {10.1088/1361-6501/acf7da},
  langid = {english}
}

@article{yuFederatedLearningAlgorithm2023,
  title = {A Federated Learning Algorithm Using Parallel-Ensemble Method on Non-{{IID}} Datasets},
  author = {Yu, Haoran and Wu, Chang and Yu, Haixin and Wei, Xuelin and Liu, Siyan and Zhang, Ying},
  year = {2023},
  month = dec,
  journal = {Complex \& Intelligent Systems},
  volume = {9},
  number = {6},
  pages = {6891--6903},
  issn = {2198-6053},
  doi = {10.1007/s40747-023-01110-7},
  langid = {english}
}

@article{zhangSpatialTemporalFederated2025,
  title = {Spatial--{{Temporal Federated Learning}} for {{Lifelong Person Re-Identification}} on {{Distributed Edges}}},
  author = {Zhang, Lei and Gao, Guanyu and Zhang, Huaizheng},
  year = {2025},
  month = feb,
  journal = {IEEE Transactions on Circuits and Systems for Video Technology},
  volume = {35},
  number = {2},
  pages = {1884--1896},
  issn = {1558-2205},
  doi = {10.1109/TCSVT.2023.3281983}
}

@article{zhuFedWeightMitigatingCovariate2025,
  title = {{{FedWeight}}: Mitigating Covariate Shift of Federated Learning on Electronic Health Records Data through Patients Re-Weighting},
  shorttitle = {{{FedWeight}}},
  author = {Zhu, He and Bai, Jun and Li, Na and Li, Xiaoxiao and Liu, Dianbo and Buckeridge, David L. and Li, Yue},
  year = {2025},
  month = may,
  journal = {npj Digital Medicine},
  volume = {8},
  number = {1},
  pages = {286},
  publisher = {Nature Publishing Group},
  issn = {2398-6352},
  doi = {10.1038/s41746-025-01661-8},
  copyright = {2025 The Author(s)},
  langid = {english}
}

@article{agrahariConceptDriftDetection2022,
  title = {Concept {{Drift Detection}} in {{Data Stream Mining}} : {{A}} Literature Review},
  shorttitle = {Concept {{Drift Detection}} in {{Data Stream Mining}}},
  author = {Agrahari, Supriya and Singh, Anil Kumar},
  year = 2022,
  month = nov,
  journal = {Journal of King Saud University - Computer and Information Sciences},
  volume = {34},
  number = {10, Part B},
  pages = {9523--9540},
  issn = {1319-1578},
  doi = {10.1016/j.jksuci.2021.11.006},
  urldate = {2025-10-26}
}

@article{qinEvolvingDomainGeneralization2023,
  title = {Evolving {{Domain Generalization}} via {{Latent Structure-Aware Sequential Autoencoder}}},
  author = {Qin, Tiexin and Wang, Shiqi and Li, Haoliang},
  year = {2023},
  month = dec,
  journal = {IEEE Transactions on Pattern Analysis and Machine Intelligence},
  volume = {45},
  number = {12},
  pages = {14514--14527},
  issn = {1939-3539},
  doi = {10.1109/TPAMI.2023.3319984},
  urldate = {2025-10-07}
}

@article{zhangAdaO2BAdaptiveOnline2025,
  title = {{{AdaO2B}}: {{Adaptive Online}} to {{Batch Conversion}} for {{Out-of-Distribution Generalization}}},
  shorttitle = {{{AdaO2B}}},
  author = {Zhang, Xiao and Dai, Sunhao and Xu, Jun and Liu, Yong and Dong, Zhenhua},
  year = {2025},
  month = apr,
  journal = {Proceedings of the AAAI Conference on Artificial Intelligence},
  volume = {39},
  number = {21},
  pages = {22596--22604},
  issn = {2374-3468},
  doi = {10.1609/aaai.v39i21.34418},
  urldate = {2025-10-07},
  copyright = {Copyright (c) 2025 Association for the Advancement of Artificial Intelligence},
  langid = {english}
}

@inproceedings{tianModelingDistributionalUncertainty2023,
  title = {Modeling the {{Distributional Uncertainty}} for {{Salient Object Detection Models}}},
  booktitle = {Proceedings of the {{IEEE}}/{{CVF Conference}} on {{Computer Vision}} and {{Pattern Recognition}}},
  author = {Tian, Xinyu and Zhang, Jing and Xiang, Mochu and Dai, Yuchao},
  year = {2023},
  pages = {19660--19670},
  urldate = {2025-10-07},
  langid = {english}
}

@article{wangEmbeddingTrajectoryOutofDistribution2024,
  title = {Embedding {{Trajectory}} for {{Out-of-Distribution Detection}} in {{Mathematical Reasoning}}},
  author = {Wang, Yiming and Zhang, Pei and Yang, Baosong and Wong, Derek F. and Zhang, Zhuosheng and Wang, Rui},
  year = {2024},
  month = dec,
  journal = {Advances in Neural Information Processing Systems},
  volume = {37},
  pages = {42965--42999},
  urldate = {2025-10-07},
  langid = {english}
}

@article{zhouFedGOGFederatedGraph2025,
  title = {{{FedGOG}}: {{Federated Graph Out-of-Distribution Generalization}} with {{Diffusion Data Exploration}} and {{Latent Embedding Decorrelation}}},
  shorttitle = {{{FedGOG}}},
  author = {Zhou, Pengyang and Chen, Chaochao and Liu, Weiming and Liao, Xinting and Shen, Wenkai and Xu, Jiahe and Fu, Zhihui and Wang, Jun and Wen, Wu and Zheng, Xiaolin},
  year = {2025},
  month = apr,
  journal = {Proceedings of the AAAI Conference on Artificial Intelligence},
  volume = {39},
  number = {21},
  pages = {22965--22973},
  issn = {2374-3468},
  doi = {10.1609/aaai.v39i21.34459},
  urldate = {2025-10-07},
  copyright = {Copyright (c) 2025 Association for the Advancement of Artificial Intelligence},
  langid = {english}
}

@article{amadorcoelhoConceptDriftDetection2023,
  title = {Concept Drift Detection with Quadtree-Based Spatial Mapping of Streaming Data},
  author = {Amador Coelho, Rodrigo and Bambirra Torres, Luiz Carlos and {Leite de Castro}, Cristiano},
  year = 2023,
  month = may,
  journal = {Information Sciences},
  volume = {625},
  pages = {578--592},
  issn = {0020-0255},
  doi = {10.1016/j.ins.2022.12.085},
  urldate = {2025-10-25}
}

@inproceedings{yuRealTimeDecisionMaking2020,
  title = {Real-{{Time Decision Making}} for {{Train Carriage Load Prediction}} via {{Multi-stream Learning}}},
  booktitle = {{{AI}} 2020: {{Advances}} in {{Artificial Intelligence}}},
  author = {Yu, Hang and Liu, Anjin and Wang, Bin and Li, Ruimin and Zhang, Guangquan and Lu, Jie},
  editor = {Gallagher, Marcus and Moustafa, Nour and Lakshika, Erandi},
  year = 2020,
  pages = {29--41},
  publisher = {Springer International Publishing},
  address = {Cham},
  doi = {10.1007/978-3-030-64984-5_3},
  isbn = {978-3-030-64984-5},
  langid = {english}
}

@article{yuRealTimePredictionSystem2022,
  title = {Real-{{Time Prediction System}} of {{Train Carriage Load Based}} on {{Multi-Stream Fuzzy Learning}}},
  author = {Yu, Hang and Lu, Jie and Liu, Anjin and Wang, Bin and Li, Ruimin and Zhang, Guangquan},
  year = 2022,
  month = sep,
  journal = {IEEE Transactions on Intelligent Transportation Systems},
  volume = {23},
  number = {9},
  pages = {15155--15165},
  issn = {1558-0016},
  doi = {10.1109/TITS.2021.3137446},
  urldate = {2026-03-07}
}

@article{stacke2020measuring,
  title={Measuring domain shift for deep learning in histopathology},
  author={Stacke, Karin and Eilertsen, Gabriel and Unger, Jonas and Lundstr{\"o}m, Claes},
  journal={IEEE journal of biomedical and health informatics},
  volume={25},
  number={2},
  pages={325--336},
  year={2020},
  publisher={IEEE}
}

@article{thiringer2026scanner,
  title={Scanner-Induced Domain Shifts Undermine the Robustness of Pathology Foundation Models},
  author={Thiringer, Erik and Gustafsson, Fredrik K and Eriksson, Kajsa Ledesma and Rantalainen, Mattias},
  journal={arXiv preprint arXiv:2601.04163},
  year={2026}
}

@inproceedings{noori2026histopath,
  title={Histopath-C: Towards Realistic Domain Shifts for Histopathology Vision-Language Adaptation},
  author={Noori, Mehrdad and Hakim, Gustavo A Vargas and Osowiechi, David and Shakeri, Fereshteh and Bahri, Ali and Yazdanpanah, Moslem and Dastani, Sahar and Ben Ayed, Ismail and Desrosiers, Christian},
  booktitle={Proceedings of the IEEE/CVF Winter Conference on Applications of Computer Vision},
  pages={4890--4900},
  year={2026}
}

@article{guan2026detecting,
  title={Detecting Performance Degradation under Data Shift in Pathology Vision-Language Model},
  author={Guan, Hao and Zhou, Li},
  journal={arXiv preprint arXiv:2601.00716},
  year={2026}
}

@inproceedings{kumariattention,
  title={Attention-based Generative Latent Replay: A Continual Learning Approach for WSI Analysis},
  author={Kumari, Pratibha and Reisenb{\"u}chler, Daniel and Bozorgpour, Afshin and Schaadt, Nadine S and Feuerhake, Friedrich and Merhof, Dorit},
  booktitle={MICCAI Workshop on Computational Pathology with Multimodal Data (COMPAYL)},
  year={2026}
}

@inproceedings{sun2022shift,
  title={SHIFT: a synthetic driving dataset for continuous multi-task domain adaptation},
  author={Sun, Tao and Segu, Mattia and Postels, Janis and Wang, Yuxuan and Van Gool, Luc and Schiele, Bernt and Tombari, Federico and Yu, Fisher},
  booktitle={Proceedings of the IEEE/CVF conference on computer vision and pattern recognition},
  pages={21371--21382},
  year={2022}
}

@article{li2024domain,
  title={Domain adaptation based object detection for autonomous driving in foggy and rainy weather},
  author={Li, Jinlong and Xu, Runsheng and Liu, Xinyu and Ma, Jin and Li, Baolu and Zou, Qin and Ma, Jiaqi and Yu, Hongkai},
  journal={IEEE Transactions on Intelligent Vehicles},
  year={2024},
  publisher={IEEE}
}

@article{xu2026vgas,
  title={VGAS: Value-Guided Action-Chunk Selection for Few-Shot Vision-Language-Action Adaptation},
  author={Xu, Changhua and Lu, Jie and Xuan, Junyu and Yu, En},
  journal={arXiv preprint arXiv:2602.07399},
  year={2026}
}

@article{gu2025climb,
  title={Climb-Odom: A robust and low-drift RGB-D inertial odometry with surface continuity constraints for climbing robots on freeform surface},
  author={Gu, Zhenfeng and Gong, Zeyu and Tan, Ke and Shi, Ying and Wu, Chong and Tao, Bo and Ding, Han},
  journal={Information Fusion},
  volume={117},
  pages={102880},
  year={2025},
  publisher={Elsevier}
}

@article{lu2025genomics,
  title={Genomics-Enhanced Cancer Risk Prediction for Personalized LLM-Driven Healthcare Recommender Systems},
  author={Lu, Kezhi and Lu, Jie and Xu, Hanshi and Guo, Kairui and Zhang, Qian and Lin, Hua and Grosser, Mark and Zhang, Yi and Zhang, Guangquan},
  journal={ACM Transactions on Information Systems},
  volume={43},
  number={6},
  pages={1--30},
  year={2025},
  publisher={ACM New York, NY}
}

@article{zeng2025sharpness,
  title={Sharpness-aware cross-domain recommendation to cold-start users},
  author={Zeng, Guohang and Zhang, Qian and Zhang, Guangquan and Lu, Jie},
  journal={IEEE Transactions on Systems, Man, and Cybernetics: Systems},
  year={2025},
  publisher={IEEE}
}

@article{lu2024amt,
  title={AMT-CDR: A deep adversarial multi-channel transfer network for cross-domain recommendation},
  author={Lu, Kezhi and Zhang, Qian and Hughes, Danny and Zhang, Guangquan and Lu, Jie},
  journal={ACM Transactions on Intelligent Systems and Technology},
  volume={15},
  number={4},
  pages={1--26},
  year={2024},
  publisher={ACM New York, NY}
}

@inproceedings{zeng2025we,
  title={Are We Really Making Recommendations Robust? Revisiting Model Evaluation for Denoising Recommendation},
  author={Zeng, Guohang and Lu, Jie and Zhang, Guangquan},
  booktitle={Proceedings of the Nineteenth ACM Conference on Recommender Systems},
  pages={706--715},
  year={2025}
}

@inproceedings{ye2026pars,
  title={PARS: Partial-Label-Learning-inspired Recommender Systems},
  author={Ye, Shanshan and Lu, Kezhi and Zhang, Guangquan and Lu, Jie},
  booktitle={Proceedings of the AAAI Conference on Artificial Intelligence},
  volume={40},
  number={33},
  pages={27800--27808},
  year={2026}
}

@article{zeng2025rosilc,
  title={RoSiLC-RS: A Robust Similar Legal Case Recommender System empowered by large language model and step-back prompting},
  author={Zeng, Guohang and Tian, George and Zhang, Guangquan and Lu, Jie},
  journal={Neurocomputing},
  volume={648},
  pages={130660},
  year={2025},
  publisher={Elsevier}
}

% \newpage


\end{document}